\documentclass[journal]{IEEEtran}
\usepackage{amsmath,amssymb,booktabs,multirow,graphicx}
\usepackage{tikz}
\usepackage{subfig}
\usetikzlibrary{arrows.meta}
\usepackage{cite}
\usepackage{url}
\newtheorem{theorem}{Theorem}
\newtheorem{proposition}{Proposition}
\newtheorem{corollary}{Corollary}
\newtheorem{remark}{Remark}
\begin{document}

\title{Correctness Metric Matters: NWD-Based Calibration Loss for Tiny Person Detection in UAV Infrared Imagery}

\author{
\IEEEauthorblockN{Tian Zhang, \emph{Member, IEEE}, and Zhirong Su}
\thanks{T. Zhang and Z. Su are with Shandong Management University, Jinan 250357, China. T. Zhang was a visiting scholar with Tsinghua University, Beijing 100084, China, during this work. (e-mails: tianzhang@sdnu.edu.cn, suzhirong007@163.com)}
}

\maketitle

\begin{abstract}
Detection calibration aligns predicted confidence with empirical correctness. Existing calibration losses inherit their correctness definition from box regression, specifically IoU against a hard threshold. We show that this target is geometrically ill-posed for tiny objects in UAV infrared imagery, where persons occupy less than 0.1\% of the image. The same pixel error is penalized more heavily as the target shrinks, so binary prediction calibration (BPC) actively suppresses confidences that should be high. On HIT-UAV, BPC with an IoU target degrades image-level localization-aware ECE (LaECE) from 0.0689 to 0.0761, 0.0649 to 0.0733, and 0.0653 to 0.0682 across Low, Mid, and High signal-to-clutter ratio (SCR) groups. We replace the IoU target with a continuous, scale-invariant normalized Wasserstein distance (NWD). The NWD target reduces LaECE\(_{\mathrm{NWD}}\) relative to BPC+IoU on all three groups (\(p<10^{-7}\)) with mAP neutral (0.9356 vs.\ 0.9347), and an independent center-offset anchor confirms that the gain is a genuine geometric correction. The improvement holds across a broad plateau of calibration weights. The NWD formulation requires no inference-time mapping layer. If a validation-fitted isotonic map is applied on top of it, calibration is lowered further, and the two corrections act on different components of the same error. The metric-level remedy has a measurable boundary. Target-background thermal pairing in the training distribution and detection-level aggregation under per-image detection-count heterogeneity lie outside it. Both are quantified independently of the NWD target. 
\end{abstract}


\section{Introduction}

Infrared person detection from unmanned aerial vehicles (UAVs) operates under a narrow set of conditions. The target is extremely small, with a mean area ratio of 0.09\%. The local signal-to-clutter ratio (SCR) is low, with medians between 1.2 and 2.6. The background carries strong thermal structure from pavement, vehicles, and vegetation. Under such conditions a detector's confidence is not merely a ranking score. Downstream decisions in search-and-rescue triage or surveillance alerting depend on whether a confidence of 0.8 corresponds to a correct detection eight times out of ten.

Calibration of object detectors has been studied mostly on general-purpose benchmarks where targets occupy a substantial fraction of the image. The dominant training-time remedy is the binary prediction calibration (BPC) loss and its variants, which penalize the gap between predicted confidence and a binary correctness label. That label is defined by intersection over union (IoU)  against a fixed threshold, typically 0.5. The definition is inherited from the regression objective and is rarely questioned.

For tiny targets it does not hold. On the HIT-UAV \cite{hit} benchmark we measure that among detections with IoU below 0.5 and no competing ground truth, 83.8\% are judged correct by normalized Wasserstein distance (NWD). Their median center offset is 4.5 pixels against a ground-truth diagonal of 22 pixels. These detections sit visually on the target. IoU penalizes them because a two-pixel shift on a $4\times12$ box produces a large relative overlap change. A calibration loss that labels them incorrect pushes their confidence down, and the resulting model is under-confident precisely where confidence should be high.

Our prior work~\cite{thermosc} established that the pairing cost survives post-hoc temperature scaling on vehicle-mounted benchmarks. We investigate whether this cost persists in the air-to-ground regime, where targets are an order of magnitude smaller and thermal backgrounds are more prominent. To address this, we focus on the correctness metric itself. We identify target scale as the geometric root cause of the IoU limitation. The pathology propagates most visibly in low SCR backgrounds, where the absolute calibration error is highest, and NWD recovers the most there. We trace the calibration cost to two sources beyond the metric. 
The main contributions are as follows.
\begin{enumerate}
\item We diagnose a structural flaw in the correctness target of BPC: the IoU link function lacks a size-independent floor, such that a fixed pixel displacement collapses IoU toward zero for tiny targets. On HIT-UAV, BPC with an IoU target raises LaECE above the uncalibrated model in every SCR group. This pathology manifests as the highest absolute error in the low-SCR group, the largest IoU-induced increment in the mid-SCR group, and the largest NWD recovery in the low-SCR group.

\item We redefine the calibration target as a continuous NWD score. Relative to BPC+IoU, the NWD target lowers LaECE across all SCR groups with mAP neutral on HIT-UAV. Physical localization is restored under an independent center-offset anchor. 

\item We identify and quantify two orthogonal calibration error sources that lie outside the metric itself. Preserving target-background thermal pairing in the training distribution raises LaECE by 0.023 in the low-SCR group without degrading mAP, and a data-side diagnostic intervention (block remap) confirms that this cost persists across the 50- and 300-epoch budgets. Detection-level aggregation bias under per-image detection-count heterogeneity decomposes into between-group and within-group components. Neither source is addressed by the NWD remedy.
\item The NWD target and post-hoc isotonic calibration are complementary. The calibration error decomposes into a geometric component that NWD reduces and a residual component that post-hoc calibration addresses, and combining the two further lowers the error.
\end{enumerate}
\section{Related Work}

\subsection{Person Detection in UAV Infrared Imagery}
Aerial person detection differs fundamentally from street-level scenarios since targets occupy only a few tens of pixels or fewer, falling into the tiny-object regime. A comprehensive survey on small object detection identifies the extreme scarcity of pixel-level information as the primary bottleneck, leading to insufficient feature representation and degraded detection accuracy~\cite{apd_survey}. Thermal infrared (TIR) imaging is the modality of choice under low illumination, and single-stage detectors dominate practical UAV deployments. Benchmarks such as HIT-UAV provide annotated TIR imagery across flight altitudes of 60--130\,m and view angles of $30^\circ$--$90^\circ$. Recent detectors on this benchmark improve tiny-target recall through feature aggregation and scale-aware assignment~\cite{ipedet,uniscale}. These works target detection accuracy. The reliability of their confidence is not evaluated. Post-hoc calibration, such as temperature scaling, is an alternative paradigm that rescales predictions after training. We avoid this route inasmuch as our prior attribution study~\cite{thermosc} demonstrated that the calibration cost originates during training and cannot be corrected by post-hoc rescaling. The present work therefore intervenes in the training objective and evaluates calibration at the image level (Section~\ref{sec:aggregation}).

\subsection{Calibration of Object Detectors}
Training-time calibration methods incorporate a calibration term into the detection objective. BPC defines the target as a binary indicator of IoU exceeding a threshold~\cite{bpc}. Cal-DETR introduces a calibrated detection transformer~\cite{caldetr}. The evaluation of these methods has recently been revisited.Kuzucu \emph{et al.} identify structural drawbacks in the prevailing combination of D-ECE and AP, including mismatched detection sets, threshold sensitivity, and neglected false negatives. They further demonstrate that under a principled evaluation protocol, post-hoc isotonic regression outperforms Cal-DETR by more than 7 D-ECE on COCO~\cite{eccv}. This implies that train-time calibration remains meaningful only if it targets a property that post-hoc rescaling cannot recover. We target the correctness label itself; post-hoc calibration merely rescales fixed predictions, whereas a mis-specified label corrupts the training signal before any rescaling can act. Unlike vehicle-mounted scenarios where IoU remains stable, the UAV regime pushes targets into the extreme tiny-object range, creating a distinct geometric pathology where the IoU link function itself fundamentally fails. We address this root cause by formulating a scale-invariant NWD target, and further expose an aggregation bias in detection-level aggregation that distorts such degradation. 

\subsection{Normalized Wasserstein Distance for Tiny Objects}
NWD models bounding boxes as two-dimensional Gaussians and measures their 2-Wasserstein distance, yielding a smooth similarity that does not collapse under small pixel shifts~\cite{nwd}. On HIT-UAV, the IoU series has been diagnosed as too sensitive to targets occupying small pixels, and NWD has been substituted for IoU in label assignment, NMS, and regression loss~\cite{nwd_hit_uav}, and utilized to stabilize bounding-box regression for infrared small targets~\cite{eflnet}. These substitutions target localization tasks, namely positive anchor assignment, NMS filtering, and regression gradient scaling. Meanwhile, using NWD as a soft assignment signal uniformly across heterogeneous target sizes can loosen supervision for larger objects and erode their localization~\cite{coswa}. Instance-level recalibration of the training signal has also been explored for small objects~\cite{semdiff}, with the calibration applied to the localization branch. Our NWD integration shifts the focus from the localization pipeline to the calibration target. Furthermore, we identify background complexity as the macroscopic manifestation that amplifies the effect of target scale (Section~\ref{sec:nwd}).
\begin{figure*}[t]
\centering
\subfloat[The matching step is shared, the link function is the only
  quantity that changes between the two conditions, and both targets
  enter the same calibration loss.\label{fig:link-a}]{%
  \includegraphics[width=0.45\linewidth]{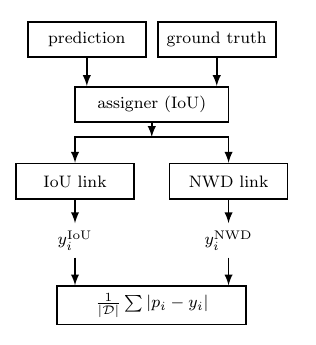}}%
\hfill
\subfloat[On a \(4\times12\) ground-truth box, a prediction displaced by
  \((d_x, d_y)=(2, 1)\) pixels receives \(\mathrm{IoU}=0.30\) and \(\mathrm{NWD}=0.87\);
  the two link functions assign different correctness to the same
  detection.\label{fig:link-b}]{%
  \includegraphics[width=0.40\linewidth]{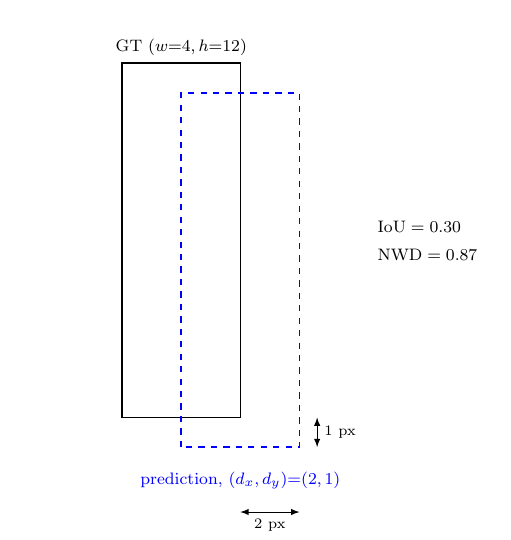}}%
\caption{The correctness link in a train-time calibration loss.}
\label{fig:link}
\end{figure*}
\section{Method}
\subsection{Problem Formulation}
\label{sec:link}

Let $p_i\in[0,1]$ be the confidence of detection $i$ and $y_i\in[0,1]$ its correctness target. A train-time calibration loss takes the form
\begin{equation}
\mathcal{L}_{\text{cal}} = \frac{1}{|\mathcal{D}|}\sum_{i\in\mathcal{D}} \left| p_i - y_i \right|.
\label{eq:cal}
\end{equation}
with $\mathcal{D}$ being the set of all predictions and $y_i=0$ for unmatched ones. The total objective is
\begin{equation}
\mathcal{L} = \mathcal{L}_{\text{det}} + \lambda \mathcal{L}_{\text{cal}},
\label{eq:total}
\end{equation}
where $\mathcal{L}_{\text{det}}$ is the detector's standard training loss, evaluated on the assigned anchors, and $\lambda$ is the calibration weight. In our experiments with YOLOv8-s, $\mathcal{L}_{\text{det}}$ comprises the classification, box regression, and distribution focal loss terms.
The correctness target is a link function \(L(b_i,\hat{b}_i)\) on the pair of boxes. IoU, its thresholded indicator, and NWD are all instantiations of \(L\)~\cite{dece,eccv,bpc}. BPC as originally formulated in~\cite{bpc} uses a hard binary IoU indicator as the correctness target. Our formulation differs in two critical respects: (i) the target is a continuous NWD score, and (ii) NWD replaces only the correctness target in the calibration loss, leaving the assigner intact. The latter distinction is essential. Table~\ref{tab:matching} demonstrates that extending NWD into the assigner degrades detection and IoU-consistent calibration. The YOLOv8 assigner relies on a soft top-k alignment metric, not a hard IoU threshold. This allows predictions with low complete IoU (CIoU) but high classification scores to be retained as matched positives, providing the exact entry point where NWD replaces the target regardless of their CIoU value. Predictions falling outside the top-k remain unmatched with \(y_i = 0\). On the diagnostic split, 83.8\% of matched positives with \(\mathrm{IoU}<0.5\) receive a higher NWD target, directly correcting their suppressed confidence. Fig.~\ref{fig:link-a} illustrates this shared matching and divergent target assignment. For the calibration loss (1) to remain informative on tiny targets, \(L\) must provide a size-independent floor under a fixed pixel displacement and must remain continuous across box parameters.  A discontinuous \(L\) would otherwise produce a training target that jumps between discrete values.

\subsection{Pairing and the Training Conditional}
\label{sec:pairing_theory}
Let $x_t$ be the target patch and $x_b$ its surrounding background. A detector trained on paired data learns $P(y=1\mid x_t,x_b)$ and evaluates this conditional on whatever background appears at test time. A detector trained on broken pairing learns $P(y=1\mid x_t)$ with $x_b$ marginalized out. The two conditionals agree when $x_b$ carries no label-relevant information and differ when it does.

In infrared imagery, pavement and vehicle residual heat produce local thermal structure that strongly resembles a pedestrian at low resolution. Under the assumption that this thermal structure is positively correlated with the label \(y\), the conditional probability satisfies \(P(y=1\mid x_t,x_b) > P(y=1\mid x_t)\) for the same target in a high-SCR background. This assumption is supported empirically by the transplant experiment in Section~\ref{sec:pairing}, where exchanging backgrounds changes the detector's confidence in the predicted direction. The difference between these two conditionals on a common test set provides one mechanism through which the calibration cost measured in Section~\ref{sec:pairing} can arise. The transplant experiment in the same section isolates the inference-time component by holding $x_t$ fixed and changing $x_b$.

\subsection{The Correctness Link: IoU Pathology, NWD Remedy, and Soft Targets}
\label{sec:correctness_link}

\subsubsection*{IoU as a link function}
Consider a prediction $b=(c_x,c_y,w,h)$ and a ground truth $\hat{b}=(\hat{c}_x,\hat{c}_y,\hat{w},\hat{h})$. Write $d_x=|c_x-\hat{c}_x|$ and $d_y=|c_y-\hat{c}_y|$. For boxes of identical size $w=\hat{w}$, $h=\hat{h}$, the intersection over union reduces to
\begin{equation}
\mathrm{IoU}(d_x,d_y)=\frac{(w-d_x)(h-d_y)}{wh+wd_y+hd_x-d_xd_y},
\label{eq:iou}
\end{equation}
for $d_x<w$, $d_y<h$, and zero otherwise.

\begin{proposition}[IoU has no scale-invariant floor]
\label{prop:iou_no_floor}
For any displacement $d_x>0$ and any $\epsilon>0$, there exists a box width $w$ such that $\mathrm{IoU}(d_x,0)<\epsilon$. The critical width is $w<d_x(1+\epsilon)/(1-\epsilon)$. No function $\ell(d_x)>0$ independent of $w$ lower-bounds IoU under horizontal displacement.
\end{proposition}

\begin{IEEEproof}
Set $d_y=0$ and $w=\hat{w}$, $h=\hat{h}$. Equation~\eqref{eq:iou} becomes $(w-d_x)/(w+d_x)$. Solving $(w-d_x)/(w+d_x)<\epsilon$ for $w$ gives $w<d_x(1+\epsilon)/(1-\epsilon)$. For fixed $d_x$, this ratio tends to zero as $w\to d_x^+$, so no positive lower bound independent of $w$ exists.
\end{IEEEproof}

The closed form \(w_c = d_x(1+\epsilon)/(1-\epsilon)\) makes the regime explicit: for a two-pixel displacement and \(\epsilon=0.5\), any target narrower than 6 px receives an IoU below 0.5 regardless of how well it is centered. This threshold is not an extreme of the HIT-UAV size distribution: it lies inside it. The median pedestrian occupies a \(12\times19\) box (median diagonal 22.5 px); under the same two-pixel displacement its IoU is approximately \(0.65\). The lower quartile, width 9 px, falls to approximately \(0.57\). The smallest pedestrians, occupying roughly \(4\times12\) boxes, fall to approximately \(0.30\). As the target shrinks, a fixed pixel error is penalized more heavily at every scale, and near the conventional 0.5 threshold the assigned correctness becomes unstable with respect to sub-pixel motion.

Beyond this size dependence, a second failure occurs when $d_x\ge w$. The intersection vanishes, yielding $\mathrm{IoU}=0$ with zero gradient in $d_x$. On a four-pixel-wide box this happens after two pixels. A prediction displaced three pixels and one displaced thirty pixels receive the same correctness and the same zero gradient from the calibration term. The link function cannot distinguish near-miss from far-miss. Both failures are inherent to the link function. Neither can be repaired by changing the loss form or the training schedule.

\subsubsection*{NWD as a link function}
NWD replaces the intersection-over-union with a distance between Gaussian embeddings. A box $b=(c_x,c_y,w,h)$ is represented as a two-dimensional Gaussian $\mathcal{N}(\mu,\Sigma)$ with
\begin{equation}
\mu=(c_x,c_y)^{\top},\qquad \Sigma=\mathrm{diag}\!\left(\frac{w^2}{4},\frac{h^2}{4}\right).
\label{eq:gauss}
\end{equation}
The standard deviation $\sigma_x=w/2$ places the box boundary at two standard deviations from the centre, where the Gaussian density has decayed to $e^{-2}\approx 0.135$ of its peak. The embedding encodes a centre-weighted prior over the box interior, in which the probability that a pixel belongs to the target decreases from the centre toward the edge. This prior matches the structure of an infrared pedestrian at low resolution, where the torso carries the thermal signature and the boundary pixels are dominated by background.
The second-order Wasserstein distance between prediction and ground truth is
\begin{equation}
W_2^2 = \|\mu_p-\mu_g\|_2^2 + \left\|\Sigma_p^{1/2}-\Sigma_g^{1/2}\right\|_F^2,
\label{eq:w2}
\end{equation}
which for diagonal covariances reduces to
\begin{equation}
W_2^2 = d_x^2 + d_y^2 + \left(\frac{w-\hat{w}}{2}\right)^2 + \left(\frac{h-\hat{h}}{2}\right)^2.
\label{eq:w2_simplified}
\end{equation}
NWD is the exponential normalization $\mathrm{NWD}(b,\hat{b})=\exp(-\sqrt{W_2^2}/C)$, with $C$ a scale constant set to the mean $\sqrt{wh}$ of ground-truth boxes in the training split. For HIT-UAV this gives $C=15.60$ pixels.

\begin{proposition}[NWD admits a scale-independent floor for a fixed dataset-level \(C\)]
\label{prop:nwd_floor}
For a fixed dataset-level scale constant \(C>0\), and for equally sized boxes with center displacement $\delta=(\delta_x,\delta_y)$, $\mathrm{NWD}(b,\hat{b})=\exp(-\|\delta\|_2/C)$ is independent of \(w\) and \(h\). For size errors bounded by $|\Delta w|,|\Delta h|\le\epsilon_s$, $\mathrm{NWD}\ge\exp(-\sqrt{\|\delta\|_2^2+\epsilon_s^2/2}/C)$.
\end{proposition}

\begin{IEEEproof}
The constant \(C\) does not depend on \(w\) or \(h\). With $w=\hat{w}$ and $h=\hat{h}$ the size terms in \eqref{eq:w2_simplified} vanish and $W_2^2=\|\delta\|_2^2$. Substituting into the NWD definition gives the first claim. For bounded size error, $W_2^2=\|\delta\|_2^2+(\Delta w/2)^2+(\Delta h/2)^2\le\|\delta\|_2^2+\epsilon_s^2/2$. Since $\exp(-\cdot/C)$ is decreasing, the lower bound follows.
\end{IEEEproof}

Proposition~\ref{prop:nwd_floor} and Proposition~\ref{prop:iou_no_floor} are two halves of the same argument. Under a fixed two-pixel displacement, IoU falls from 0.88 to 0.30 as the box shrinks from \(32\times96\) to \(4\times12\); NWD stays near 0.87. The calibration target should not penalize a fixed pixel error more heavily merely because the object is small, and only one of the two link functions satisfies this requirement. Because \(C\) is dataset-level, the floor is anchored to one characteristic scale. On FLIR, where target scales span a wider range, this predicts a weaker signal for large targets—which is what Section~\ref{sec:flir} observes.

Instead of modifying the IoU penalty term (as in GIoU~\cite{giou} or DIoU/CIoU~\cite{diou}), our approach replaces the intersection-over-union paradigm entirely. These IoU variants remain bounded by the same scale-dependent floor (Proposition~\ref{prop:iou_no_floor}), whereas NWD introduces a fundamentally different Gaussian embedding that guarantees scale invariance for a fixed scale constant \(C\) (Proposition~\ref{prop:nwd_floor}).

The two propositions place IoU and NWD on opposite sides of the property that matters for calibration. IoU has no floor under fixed displacement (Proposition~\ref{prop:iou_no_floor}). NWD has one, and the floor depends only on the displacement and $C$ (Proposition~\ref{prop:nwd_floor}). At the operating point of interest, $d_x=2$, $d_y=1$, $w=\hat{w}$, $h=\hat{h}$, Proposition~\ref{prop:nwd_floor} gives $\|\delta\|_2=\sqrt{5}\approx 2.24$ and $\mathrm{NWD}\approx\exp(-2.24/15.60)\approx 0.87$. The same displacement gives $\mathrm{IoU}\approx 0.30$. Fig.~\ref{fig:link-b} visualizes this geometric discrepancy on a $4\times 12$ box. The two link functions disagree on the correctness of this detection by $0.57$.

The disagreement is size-dependent, and this is the property the calibration target requires. For the $32\times96$ box, NWD yields $0.87$, identical to IoU within numerical precision. For the $4\times12$ box, NWD remains at $0.87$ while IoU drops to $0.30$. IoU penalizes the same pixel error more heavily as the target shrinks, because the intersection term $(w-d_x)(h-d_y)$ loses a larger fraction of the box area. NWD does not, with its center term $d_x^2+d_y^2$ in \eqref{eq:w2_simplified} independent of $w$ and $h$. A fixed pixel error is penalized equally regardless of target width, and the calibration target should reflect that.

\begin{figure*}[t]
\centering
\subfloat[]{\includegraphics[width=0.32\textwidth]{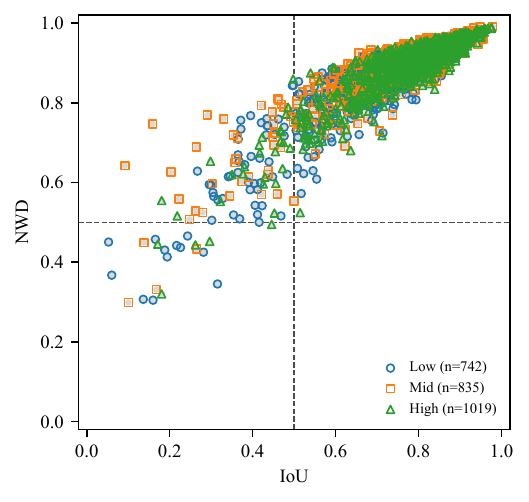}}\hfill
\subfloat[]{\includegraphics[width=0.32\textwidth]{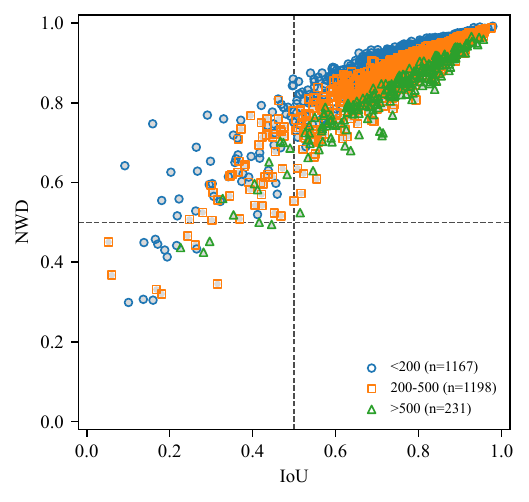}}\hfill
\subfloat[]{\includegraphics[width=0.32\textwidth]{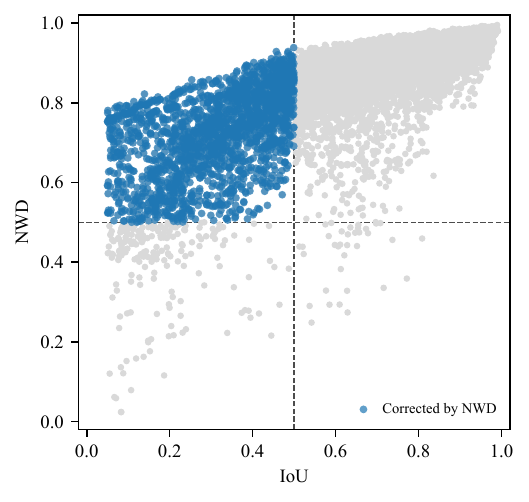}}
\caption{NWD against IoU for matched true positives. Dashed lines mark the conventional 0.5 threshold on each axis. (a) HIT-UAV points colored by SCR group. (b) HIT-UAV points colored by ground-truth area in pixels. (c) FLIR ADAS\cite{flir_dataset} points; blue markers highlight detections that IoU labels incorrect (IoU $<$ 0.5) but NWD labels correct (NWD $>$ 0.5). The lower-right quadrant is empty across all panels. In HIT-UAV, 83.8\% of detections fall into the upper-left quadrant; on FLIR, the correction rate is 90.5\%.}
\label{fig:scatter}
\end{figure*}

The disagreement is also one-sided. On the diagnostic split, $83.8\%$ of detections with $\mathrm{IoU}<0.5$ and no competing ground truth receive $\mathrm{NWD}>0.5$. The reverse direction does not occur: no detection in the diagnostic set has $\mathrm{IoU}\ge 0.5$ and $\mathrm{NWD}<0.5$. The set of detections on which the two link functions disagree lies entirely on the side that Proposition~\ref{prop:iou_no_floor} predicts. Fig.~\ref{fig:scatter}(a) visualizes this one-sided disagreement on the diagnostic split. To ensure that our visual correctness proxy \(y^\ast = \mathbb{I}[\mathrm{NWD} > \tau]\) is not sensitive to the threshold choice, we evaluate the one-sided disagreement at \(\tau \in \{0.3, 0.5, 0.7\}\) on the diagnostic split. The proportion of detections with \(\mathrm{IoU} < 0.5\) that are judged correct by NWD is \(99.2\%\) at \(\tau=0.3\) and \(83.8\%\) at \(\tau=0.5\). The reverse case (detections with \(\mathrm{IoU} \ge 0.5\) but \(\mathrm{NWD} < \tau\)) never occurs at these thresholds (0.0\%). IoU therefore penalizes a large portion of visually correct detections, while NWD never penalizes those that IoU considers correct. At \(\tau=0.7\), the reverse false negative rate rises to \(14.6\%\), because a higher threshold makes NWD stricter and causes it to reject some detections that IoU labels correct. A threshold of 0.5 balances the two: it corrects most of IoU's misjudgments and keeps the reverse false negative rate at zero.
\subsubsection*{Soft targets}
Given a link function, the calibration loss admits a hard target $y_i=\mathbb{I}[L(b_i,\hat{b}_i)\ge\tau]$ or a soft target $y_i=L(b_i,\hat{b}_i)$. We use the soft target, and Propositions~\ref{prop:iou_no_floor} and~\ref{prop:nwd_floor} explain why the choice is not cosmetic.

With a hard target, the loss is $|p_i-y_i|$ where $y_i\in\{0,1\}$. Its gradient in $p_i$ is $\pm 1$ almost everywhere and undefined at $p_i=y_i$, whereas  the target itself jumps at $L=\tau$. A prediction with $L=0.51$ and one with $L=0.99$ receive the same target, so the loss cannot separate a detection that barely clears the threshold from one that sits on the target. Under IoU this collapses further. For a narrow box, Proposition~\ref{prop:iou_no_floor} places both $L=0.30$ and $L=0.00$ below the threshold, so they receive the same hard target of zero. The loss provides no gradient in $d_x$ to separate them. The hard target discards the floor that Proposition~\ref{prop:nwd_floor} provides.

The hard target also changes the positive-class prevalence, the fraction of matched predictions labeled positive. Thresholding NWD at \(\tau=0.5\) labels \(97.1\%\) of matched predictions as positive, against \(80.2\%\) under IoU. This gap introduces a class-imbalance confound. The soft target avoids this by construction, ensuring the calibration signal depends solely on the continuous link value.

Each summand of the calibration term is 1-Lipschitz in $p_i$. Since $y_i\in[0,1]$, the partial derivative $|\partial \mathcal{L}_{\text{cal}}/\partial p_i|$ equals $1/|\mathcal{D}|$ almost everywhere, and the overall objective is 1-Lipschitz with respect to the $\ell_\infty$ norm on the confidence vector. Increasing $\lambda$ therefore changes the relative weight of the calibration term against $\mathcal{L}_{\text{det}}$ without changing the local conditioning of the calibration term itself.

\subsection{Aggregation Bias}
\label{sec:aggregation_theory}
Calibration error in this setting stems from two independent sources. The link function dictates what the training loss treats as correct (Section~\ref{sec:correctness_link}), and the aggregation protocol governs how the error is estimated from a set of detections. Propositions~\ref{prop:iou_no_floor} and~\ref{prop:nwd_floor} address the link function, while the following proposition addresses the aggregation protocol. These two limitations operate independently. A correct link function does not remove the aggregation bias, nor can an unbiased aggregation compensate for a mis-specified metric.

Let \(I\) be the set of images and \(D_j\) the set of detections in image \(j\), with \(D=\cup_j D_j\). For notational simplicity, we note that \(D\) and \(\mathcal{D}\) (introduced in \eqref{eq:cal}) refer to the same global detection set, and the two symbols are used interchangeably. Detection-level aggregation computes $\frac{1}{|D|}\sum_{i\in D}|y_i-p_i|$. Image-level aggregation computes $\frac{1}{|I|}\sum_{j\in I}\frac{1}{|D_j|}\sum_{i\in D_j}|y_i-p_i|$, where \(I\) denotes the set of images containing at least one detection; empty images are excluded. When every image contributes the same number of detections, the two aggregations coincide; otherwise they generally differ.

The two aggregations differ in what they weight. Detection-level aggregation weights each detection equally; image-level aggregation weights each image equally. In deployment, an operator inspects images, not detections. A false alarm on a clutter-heavy image that happens to produce twenty candidate boxes should not count twenty times as much as a false alarm on a clean image with one box. The image-level estimator therefore matches the unit at which calibration matters downstream, and the detection-level estimator's deviation from it is a bias relative to that unit.

\begin{proposition}[Aggregation bias decomposition]
\label{prop:dilution}
Let \(r_k = |D\cap G_k|/|D|\) and \(s_k = |I\cap G_k|/|I|\) be the detection and image fractions of group \(k\), with \(w_k = r_k/s_k\). Let
\begin{equation}
e_{det,k} = \frac{1}{|D\cap G_k|}\sum_{i\in D\cap G_k}|y_i - p_i|,
\label{eq:edet_k}
\end{equation}
\begin{equation}
e_{img,k} = \frac{1}{|I\cap G_k|}\sum_{j\in I\cap G_k}\frac{1}{|D_j|}\sum_{i\in D_j}|y_i - p_i|
\label{eq:eimg_k}
\end{equation}
be the detection-level and image-level mean errors within group \(k\). Then
\begin{eqnarray}
\lefteqn{\mathcal{B}_{\mathrm{agg}} := \mathrm{LaECE}^{\mathrm{det}} - \mathrm{LaECE}^{\mathrm{img}}} \nonumber\\
&=& \underbrace{\sum_k s_k(w_k - 1)e_{img,k}}_{\text{Between-group}} + \underbrace{\sum_k s_k w_k (e_{det,k} - e_{img,k})}_{\text{Within-group}}.\nonumber\\
\label{eq:agg_bias}
\end{eqnarray}
The first term is a density-weighted covariance between the group density ratio \(w_k\) and the image-level mean error \(e_{img,k}\), while the second term captures the weighted mean difference between detection-level and image-level errors within each group.
\end{proposition}

\begin{IEEEproof}
By definition, \(\mathrm{LaECE}^{\mathrm{det}} = \sum_k r_k e_{det,k}\) and \(\mathrm{LaECE}^{\mathrm{img}} = \sum_k s_k e_{img,k}\). Their difference is:
\[
\sum_k r_k e_{det,k} - \sum_k s_k e_{img,k} = \sum_k s_k (w_k e_{det,k} - e_{img,k}).
\]
Adding and subtracting \(s_k w_k e_{img,k}\) yields
\begin{eqnarray}
\lefteqn{\sum_k s_k w_k e_{det,k} - \sum_k s_k w_k e_{img,k}} \nonumber \\
& & \quad + \sum_k s_k w_k e_{img,k} - \sum_k s_k e_{img,k}. \nonumber
\end{eqnarray}
Factoring gives the Between-group and Within-group components.
\end{IEEEproof}
On our data it isolates the source of the aggregation discrepancy. Of the total gap, 88\% originates from within-group count heterogeneity, not from between-group density ratios (Corollary~\ref{cor:direction}).

\begin{corollary}[Empirical dominance of the within-group count-heterogeneity term]
\label{cor:direction}
In our data, the Between-group term is \(0.00039\) and the Within-group term is \(0.00296\), meaning \(88\%\) of the aggregation gap originates from per-image detection-count heterogeneity, not from the group density ratios. This is an empirical observation, not a theoretical necessity: the Within-group term can be positive or negative in principle depending on whether detection-level errors exceed image-level errors within each group. The sign pattern is consistent across both LaECE\(_{\mathrm{NWD}}\) and LaECE\(_{\mathrm{IoU}}\), with the per-group decomposition reported in Table~\ref{tab:within_decomp}. The positive dominance is driven by the Mid and High SCR groups, where a small number of clutter-heavy images have detection counts far exceeding the group average and also carry higher-than-average errors, inflating the detection-level mean relative to the image-level mean. The Low-SCR group, despite its high total detection count and \(78.6\%\) false-positive rate, contributes negatively to the Within-group term, because its per-image detection-level mean is systematically lower than its image-level mean: the large number of low-confidence false positives within each image pulls the detection-level average down, while the image-level average is less affected because each image contributes only one value. Image-level aggregation suppresses this intra-group heterogeneity by averaging within each image before aggregating across images.
\end{corollary}
\begin{table}[t]
\centering
\caption{Per-group decomposition of the Within-group term under both metrics. Entries are \(e_{det,k} - e_{img,k}\) and its \(r_k\)-weighted contribution.}
\label{tab:within_decomp}
\setlength{\tabcolsep}{3.5pt}
\resizebox{\columnwidth}{!}{%
\begin{tabular}{lcccc}
\toprule
\multirow{2}{*}{Group} & \multicolumn{2}{c}{LaECE\(_{\mathrm{NWD}}\)} & \multicolumn{2}{c}{LaECE\(_{\mathrm{IoU}}\)} \\
\cmidrule(lr){2-3}\cmidrule(lr){4-5}
 & \(\Delta e\) & \(r_k \Delta e\) & \(\Delta e\) & \(r_k \Delta e\) \\
\midrule
Low  & \(-0.0017\) & \(-0.00074\) & \(-0.0016\) & \(-0.00072\) \\
Mid  & \(+0.0070\) & \(+0.00198\) & \(+0.0070\) & \(+0.00197\) \\
High & \(+0.0061\) & \(+0.00171\) & \(+0.0061\) & \(+0.00171\) \\
\midrule
Sum  & & \(+0.00296\) & & \(+0.00296\) \\
\bottomrule
\end{tabular}}
\end{table}

This aggregation bias is complementary to the object-level calibration error (OCE) proposed by Park~\emph{et al.}~\cite{oce}, which aggregates predictions per ground-truth object as opposed to per prediction. While OCE resolves the mismatch between prediction count and object count by redefining the aggregation unit, our Proposition~\ref{prop:dilution} quantifies the structural discrepancy between image-level and detection-level estimators. By decomposing the gap into Between-group density differences and Within-group weighted mean differences, we show that the dominance of per-image detection-count heterogeneity makes detection-level aggregation inherently unreliable. Both analyses converge on the same conclusion: the choice of aggregation unit fundamentally alters the estimated calibration error.

\subsection{Structural Bound of Calibration Error}
\label{sec:unified_bound}

To derive the structural bound, we introduce a binary intermediate variable \(y_i^\ast = \mathbb{I}[\mathrm{NWD} > 0.5]\) for matched detections (\(y_i^\ast = 0\) otherwise). This variable serves as a mathematical bridge for Theorem~\ref{thm:unified_bound} to connect the continuous link function \(L\) with the detector's confidence. The empirical gains reported in our experiments are corroborated by multiple anchors, including the independent LaECE\(_{\mathrm{IoU}}\) and Brier\(_{\mathrm{IoU}}\), and are therefore not artifacts of metric alignment.

We measure the calibration error by the per-sample image-level calibration error \(\mathcal{E}_{\mathrm{MAD}}\). Let \(p_i\) be the confidence of detection \(i\), and let \(y_i\in[0,1]\) be its continuous correctness target, with \(y_i=0\) for unmatched detections. Let \(I\) denote the set of images containing at least one detection. For each image \(j\in I\), the mean absolute deviation is
\begin{equation}
e_j = \frac{1}{|D_j|}\sum_{i\in D_j}|p_i - y_i|,
\label{eq:image_mad}
\end{equation}
and the image-level calibration error is
\begin{equation}
\mathcal{E}_{\mathrm{MAD}} = \frac{1}{|I|}\sum_{j\in I} e_j.
\label{eq:emad}
\end{equation}

We define the geometric distortion of a link function \(L\) as
\begin{equation}
\mathcal{D}_{\mathrm{geom}}(L)
:= \frac{1}{|I|}\sum_{j\in I}\frac{1}{|D_j|}\sum_{i\in D_j}
\bigl|L(b_i,\hat b_i) - y_i^\ast\bigr|,
\label{eq:geom_distortion}
\end{equation}
and the detection-level residual error as
\begin{equation}
\mathcal{E}_{\mathrm{res}}^{\mathrm{det}}
:= \frac{1}{|\mathcal{D}|}\sum_{i\in\mathcal{D}}
\bigl|y_i^\ast - p_i\bigr|,
\label{eq:residual_error}
\end{equation}
which is the detection-level LaECE computed with target \(y^\ast\).
The aggregation gap under \(y^\ast\) is defined as the difference between
the detection-level and image-level LaECE computed with target \(y^\ast\),
\begin{equation}
\mathcal{B}_{\mathrm{agg}}^\ast
:= \mathcal{E}^{\ast,\mathrm{det}} - \mathcal{E}^{\ast,\mathrm{img}},
\label{eq:agg_gap_star}
\end{equation}
which, by Proposition~\ref{prop:dilution} applied to the target \(y^\ast\),
decomposes exactly as
\begin{equation}
\mathcal{B}_{\mathrm{agg}}^\ast
= \underbrace{\sum_k s_k(w_k^\ast - 1)\bar e_k^{\ast,\mathrm{img}}}_{\mathcal{B}_{\mathrm{between}}^\ast}
+ \underbrace{\sum_k s_k w_k^\ast\,(\bar e_k^{\ast,\mathrm{det}} - \bar e_k^{\ast,\mathrm{img}})}_{\mathcal{B}_{\mathrm{within}}^\ast},
\label{eq:agg_gap_decomp}
\end{equation}
where \(w_k^\ast = r_k^\ast/s_k\), \(r_k^\ast = |D\cap G_k|/|D|\), and
\(s_k = |I\cap G_k|/|I|\). The per-group means under the proxy \(y^\ast\) are
\begin{equation}
\bar e_k^{\ast,\mathrm{det}} = \frac{1}{|D\cap G_k|}\sum_{i\in D\cap G_k}|y_i^\ast - p_i|,
\label{eq:edet_k_star}
\end{equation}
\begin{equation}
\bar e_k^{\ast,\mathrm{img}} = \frac{1}{|I\cap G_k|}\sum_{j\in I\cap G_k}\frac{1}{|D_j|}\sum_{i\in D_j}|y_i^\ast - p_i|,
\label{eq:eimg_k_star}
\end{equation}
which are the \(y^\ast\)-analogs of \eqref{eq:edet_k} and \eqref{eq:eimg_k}. The between-group term equals
\(\mathrm{Cov}_s(w_k^\ast, \bar e_k^{\ast,\mathrm{img}})\), the covariance
between group-level detection density and image-weighted within-group error.
The within-group term captures the residual gap from unequal per-image
detection counts inside each group. In the notation of
Proposition~\ref{prop:dilution}, \(\mathcal{B}_{\mathrm{agg}}^\ast\) is the
sum of both components, not the between-group covariance alone.

The following theorem formalizes these structural sources of calibration error, separating the metric distortion from the aggregation bias.
\begin{theorem}[Structural Bound of Calibration Error]
\label{thm:unified_bound}
For any link function \(L: \mathcal{B} \times \mathcal{B} \to [0,1]\) defined on matched detections, with the convention \(L(b_i,\hat{b}_i) = 0\) for unmatched detections, the per-sample image-level calibration error \(\mathcal{E}_{\mathrm{MAD}}\) satisfies
\begin{equation}
\mathcal{E}_{\mathrm{MAD}}
\;\le\;
\underbrace{\mathcal{D}_{\mathrm{geom}}(L)}_{\text{geometric distortion}}
\;+\;
\underbrace{\mathcal{E}_{\mathrm{res}}^{\mathrm{det}}}_{\text{detection-level residual}}
\;+\;
\underbrace{|\mathcal{B}_{\mathrm{agg}}^\ast|}_{\text{aggregation gap}}.
\label{eq:unified_bound}
\end{equation}
In the bound above, \(y_i = L(b_i,\hat{b}_i)\) for matched detections and \(y_i = 0\) for unmatched detections, so that \(\mathcal{E}_{\mathrm{MAD}}\) is the image-level calibration error under the continuous link function \(L\). The right side uses the binary proxy \(y^\ast\) as an intermediate anchor via the triangle inequality \(|L(b_i,\hat{b}_i) - p_i| \le |L(b_i,\hat{b}_i) - y_i^\ast| + |y_i^\ast - p_i|\).
\end{theorem}

\begin{IEEEproof}
For every detection \(i\), the triangle inequality gives
\[
\bigl|L(b_i,\hat b_i) - p_i\bigr|
\le \bigl|L(b_i,\hat b_i) - y_i^\ast\bigr|
 + \bigl|y_i^\ast - p_i\bigr|.
\]
Averaging over images with equal weight (detections within an image first
averaged, then images), we obtain
\begin{equation}
\mathcal{E}_{\mathrm{MAD}}
\;\le\; \mathcal{D}_{\mathrm{geom}}(L) + \mathcal{E}^{\ast,\mathrm{img}},
\label{eq:step1}
\end{equation}
where
\[
\mathcal{E}^{\ast,\mathrm{img}}
:= \frac{1}{|I|}\sum_{j\in I}\frac{1}{|D_j|}\sum_{i\in D_j}
\bigl|y_i^\ast - p_i\bigr|
\]
is the image-level LaECE with \(y^\ast\) as correctness target.

Apply Proposition~\ref{prop:dilution} to the correctness target \(y^\ast\). The decomposition holds for any target, so replacing \(e_{det,k}\) and \(e_{img,k}\) with their \(y^\ast\)-analogs gives
\begin{eqnarray}
\mathcal{E}^{\ast,\mathrm{det}} - \mathcal{E}^{\ast,\mathrm{img}} 
&=& \sum_k s_k(w_k^\ast - 1)\bar e_k^{\ast,\mathrm{img}} \nonumber \\
& & + \sum_k s_k w_k^\ast (\bar e_k^{\ast,\mathrm{det}} - \bar e_k^{\ast,\mathrm{img}}) = \mathcal{B}_{\mathrm{agg}}^\ast, \nonumber
\end{eqnarray}
where \(\mathcal{E}^{\ast,\mathrm{det}} = \mathcal{E}_{\mathrm{res}}^{\mathrm{det}}\). Rearranging gives \(\mathcal{E}^{\ast,\mathrm{img}} = \mathcal{E}_{\mathrm{res}}^{\mathrm{det}} - \mathcal{B}_{\mathrm{agg}}^\ast\). Substituting into \eqref{eq:step1} and bounding \(-\mathcal{B}_{\mathrm{agg}}^\ast \le |\mathcal{B}_{\mathrm{agg}}^\ast|\), we arrive at the general bound \(\mathcal{E}_{\mathrm{MAD}} \le \mathcal{D}_{\mathrm{geom}}(L) + \mathcal{E}_{\mathrm{res}}^{\mathrm{det}} + |\mathcal{B}_{\mathrm{agg}}^\ast|\).
\end{IEEEproof}

\begin{corollary}[Tightening of the bound in the present setting]
\label{cor:tight}
In our experiments, the total aggregation gap \(\mathcal{B}_{\mathrm{agg}}^\ast\) is positive, meaning the detection-level estimate overestimates the image-level error. The proof of Theorem~\ref{thm:unified_bound} then yields a tighter intermediate bound \(\mathcal{E}_{\mathrm{MAD}} \le \mathcal{D}_{\mathrm{geom}}(L) + \mathcal{E}_{\mathrm{res}}^{\mathrm{det}} - \mathcal{B}_{\mathrm{agg}}^\ast\), which is strictly smaller than the relaxed form \eqref{eq:unified_bound}. The sign condition applies to the \emph{total} gap, not to each group: Proposition~\ref{prop:dilution} shows that the within-group term can be negative in specific groups (e.g., Low-SCR in our data) while the sum remains positive. The bound therefore couples the three calibration sources, with the direction of the aggregation bias determining whether the estimate over- or under-shoots the true image-level error.
\end{corollary}

\begin{table}[t]
\centering
\caption{Decomposition of the calibration error under five conditions on the HIT-UAV test split (three seeds, mean \(\pm\) std). \(\mathcal{D}_{\mathrm{geom}}\) depends only on the link function \(L\). TS and isotonic regression leave \(L\) unchanged and share the BPC+IoU value; NWD replaces \(L\) and reduces \(\mathcal{D}_{\mathrm{geom}}\). \(\mathcal{E}_{\mathrm{MAD}}\) is computed with each condition's own \(L\) and is not comparable across rows; it is also not the group-averaged LaECE of Table~\ref{tab:main} and is not additive with it.}
\label{tab:component_decomp}
\setlength{\tabcolsep}{3.5pt}
\resizebox{\columnwidth}{!}{%
\begin{tabular}{lcccc}
\toprule
Condition & \(\mathcal{D}_{\mathrm{geom}}\) & \(\mathcal{E}_{\mathrm{res}}^{\mathrm{det}}\) & \(\mathcal{B}_{\mathrm{agg}}^\ast\) & \(\mathcal{E}_{\mathrm{MAD}}\) \\
\midrule
No BPC (50 ep) & \(0.0337\pm0.0055\) & \(0.1048\pm0.0108\) & \(+0.0242\pm0.0018\) & \(0.0498\pm0.0061\) \\
BPC+IoU ($\lambda=50$) & \(0.0432\pm0.0050\) & \(0.1168\pm0.0046\) & \(+0.0257\pm0.0022\) & \(0.0548\pm0.0030\) \\
BPC+NWD ($\lambda=50$) & \(0.0205\pm0.0022\) & \(0.1087\pm0.0040\) & \(+0.0218\pm0.0019\) & \(0.0668\pm0.0037\) \\
\quad + Temperature Scaling & \(0.0432\pm0.0050\) & \(0.1044\pm0.0049\) & \(+0.0235\pm0.0017\) & \(0.0494\pm0.0032\) \\
\quad + Fair Isotonic & \(0.0432\pm0.0050\) & \(0.1043\pm0.0071\) & \(+0.0215\pm0.0016\) & \(0.0467\pm0.0052\) \\
\bottomrule
\end{tabular}}
\end{table}

Across five conditions, the decomposition satisfies the inequality of Theorem~\ref{thm:unified_bound} in every case (Table~\ref{tab:component_decomp}).  Using the tightened bound of Corollary~\ref{cor:tight}, the minimum is 0.1074 (BPC+NWD), which exceeds the corresponding \(\mathcal{E}_{\mathrm{MAD}}\) of 0.0668.

The decomposition separates mechanisms that a single LaECE number conflates. Temperature scaling and isotonic regression both act on the confidence mapping and leave \(L\) unchanged. Their \(\mathcal{D}_{\mathrm{geom}}\) is identical to the BPC+IoU value (\(0.0432\)), because \(\mathcal{D}_{\mathrm{geom}}\) is the gap between \(L\) and the binary anchor and neither method alters \(L\). Their \(\mathcal{E}_{\mathrm{res}}^{\mathrm{det}}\) drops from \(0.1168\) to \(0.1044\) and \(0.1043\), which is where the confidence remapping acts. NWD replaces \(L\), so its \(\mathcal{D}_{\mathrm{geom}}\) drops to \(0.0205\), below the uncalibrated model's \(0.0337\). Its \(\mathcal{E}_{\mathrm{res}}^{\mathrm{det}}\) (\(0.1087\)) stays in the range set by the IoU-link conditions. Between the two correctable components, \(\mathcal{D}_{\mathrm{geom}}\) and \(\mathcal{B}_{\mathrm{agg}}^\ast\), the geometric distortion is larger by a factor of \(1.7\) in the BPC+IoU condition (\(0.0432\) against \(0.0257\)). The NWD substitution therefore addresses the dominant correctable term.

\begin{remark}[Asymptotic Saturation on Tiny Targets]
\label{rem:saturation}
The bound in \eqref{eq:unified_bound} explains the structural failure of
IoU. For a fixed pixel displacement \(d_x>0\), the lack of a scale-invariant floor (Proposition~\ref{prop:iou_no_floor}) causes IoU to collapse toward zero even for visually correct detections as the target width \(w \to d_x^+\). Restricting to the visually correct subset \(S^\ast = \{i : y_i^\ast = 1\}\), the conditional average of \(|\mathrm{IoU} - 1|\) on this subset approaches its maximum:
\[
\frac{1}{|S^\ast|}\sum_{i\in S^\ast} \bigl|\mathrm{IoU}(b_i,\hat b_i) - 1\bigr| \;\longrightarrow\; 1 \quad \text{as } w \to d_x^+.
\]
The convergence is monotone in \(w\) for fixed \(d_x\): for any \(\epsilon>0\), there exists a critical width \(w_c(\epsilon)\) such that for all \(w < w_c\), the conditional average exceeds \(1-\epsilon\).
In contrast, the scale-invariant floor of NWD
(Proposition~\ref{prop:nwd_floor}) keeps
\(\mathrm{NWD}(b_i,\hat b_i)\ge \exp(-\|\delta\|_2/C)\) on the same subset,
bounding its geometric distortion away from \(1\). This saturation of IoU
on tiny visually correct detections is consistent with the size
stratification in Table~\ref{tab:scale_scr}: the smallest size bin, where
IoU's geometric distortion is largest, exhibits the most consistent
SCR-wise improvement under NWD.
\end{remark}

\subsection{Inference and Evaluation Protocol}
\label{sec:protocol}

We confine the calibration term to the training objective in \eqref{eq:total}. At inference only $\mathcal{L}_{\text{det}}$ is relevant, and the deployed network is structurally identical to a detector trained without calibration. The exported model has no additional parameters, no extra forward pass, and no post-NMS rescaling. The calibration strength $\lambda$ is fixed at training and does not appear in the exported model.

This training-time formulation contrasts sharply with post-hoc calibration, which requires a rescaling operation after detection and must be re-estimated whenever the operating threshold or deployment distribution changes. For UAV deployment on embedded hardware, where memory and latency budgets are tight, the absence of an inference-time calibration module is a practical property of the training-time formulation.

Calibration is measured by \(\mathcal{E}_{\mathrm{MAD}}\), as defined in Section~\ref{sec:unified_bound}. In the experiments, the continuous correctness target \(y_i\) is instantiated as the matched IoU or NWD score for true positives, and \(y_i=0\) otherwise. LaECE\(_{\mathrm{IoU}}\) and LaECE\(_{\mathrm{NWD}}\) denote the group-level average of \(e_j\) under these two instantiations. Reporting both variants ensures that the observed gains are not artifacts of a specific correctness target.

\section{Experiments}

\subsection{Dataset and Implementation}
HIT-UAV contains 2,898 infrared images (2,029/290/579 train/val/test) captured at flight altitudes of 60--130\,m and view angles of $30^\circ$--$90^\circ$, with day and night sessions. We use the single-class Person setting. All detectors are YOLOv8-s trained at $640\times640$ with mosaic and flip augmentation and no color augmentation. The 300-epoch baselines and the pairing-broken model are used only for the pairing-cost experiment (Section~\ref{sec:pairing}). The calibration comparison uses a separate 50-epoch training protocol. Within this protocol, the No-BPC condition is obtained by setting \(\lambda=0\) in the same BPC training script used for the calibrated variants. It shares the identical 50-epoch schedule, augmentation, and matching strategy as BPC+IoU and BPC+NWD, and differs only in the absence of the calibration term. This provides a same-budget control for isolating the effect of the correctness target from training-duration effects. Seeds are 0, 1, and 42 throughout. SCR is computed per target as $|\mu_{\text{target}}-\mu_{\text{local bg}}|/\sigma_{\text{local bg}}$ with a 20-pixel dilation, and images are partitioned into Low/Mid/High groups at fixed tertile boundaries ($1.2859$ and $2.3601$).The validation split (290 images) is used for monitoring training convergence and, for the post-hoc calibration baselines in Table~\ref{tab:posthoc}, for fitting the calibration mappings, following standard practice in the calibration literature. All reported mAP values are on the test split (579 images) unless explicitly stated otherwise. Models trained for 300 epochs (Section~\ref{sec:pairing}) and models trained for 50 epochs (Section~\ref{sec:nwd}) are reported in separate tables and are not compared against each other on mAP. Unless otherwise stated, all reported metrics represent the mean and standard deviation over three seeds (0, 1, 42). For statistical tests, we compute the per-image mean of the LaECE across the three seeds and apply paired Wilcoxon signed-rank tests, where \(n\) denotes the number of unique test images per SCR group. For each group-level comparison, we additionally verify that the direction of the effect is consistent across all three seeds. The per-image mean across seeds is treated as a single observation per image; the Wilcoxon test operates at the image level, where images are independent samples. Holm--Bonferroni correction is applied to the three SCR-group comparisons within each table. Target scale stratification, IoU threshold stratification, C sensitivity analysis, and selective prediction are conducted on seed 42 to isolate specific mechanisms.

\subsection{Pairing Cost}
\label{sec:pairing}

Our prior work~\cite{thermosc} decoupled pairing correspondence from absolute thermal scale. Since HIT-UAV provides only 8-bit infrared images without radiometric temperature, we restrict the controlled attribution strictly to the pairing axis. We isolate this factor by comparing original images against block-remapped versions. Section~\ref{sec:pairing_theory} predicts that the two conditionals $P(y=1\mid x_t,x_b)$ and $P(y=1\mid x_t)$ differ whenever the background carries label-relevant structure. We test this prediction by constructing a training set in which $x_b$ is decorrelated from $x_t$. We construct a pairing-broken training set by partitioning each image into $32\times32$ blocks and applying an independent linear grayscale remap per block ($a\in[0.9,1.1]$, $b\in[-0.06,0.06]$), which preserves within-block ordering and the global first two moments while destroying cross-block thermal correspondence. The two models are otherwise identical.

\begin{table}[t]
\centering
\caption{Pairing cost: pairing-preserving (A) vs.\ pairing-broken (B). LaECE$_{\text{IoU}}$ is image-level, averaged over three seeds.}
\label{tab:pairing}
\begin{tabular}{lccc}
\toprule
 & A & B & $\Delta$(A$-$B) \\
\midrule
mAP$_{50}$ & 0.9284 & 0.9253 & $-0.0031$ \\
mAP$_{50\text{-}95}$ & 0.5134 & 0.5094 & $-0.0040$ \\
LaECE$_{\text{IoU}}$ Low & 0.0726 & 0.0494 & $+0.0232$ ($p=4.4\times10^{-15}$) \\
LaECE$_{\text{IoU}}$ Mid & 0.0627 & 0.0492 & $+0.0135$ ($p=1.5\times10^{-12}$) \\
LaECE$_{\text{IoU}}$ High & 0.0617 & 0.0442 & $+0.0176$ ($p=4.9\times10^{-14}$) \\
\bottomrule
\end{tabular}
\end{table}

 Pairing preservation raises LaECE\(_{\text{IoU}}\) in every group by an order of magnitude in significance, while detection accuracy differs by less than 0.005 mAP (Table \ref{tab:pairing}). The calibration cost is isolated from detection degradation.

\begin{figure}[!t]
\centering
\subfloat[Low-SCR targets]{\includegraphics[width=\columnwidth]{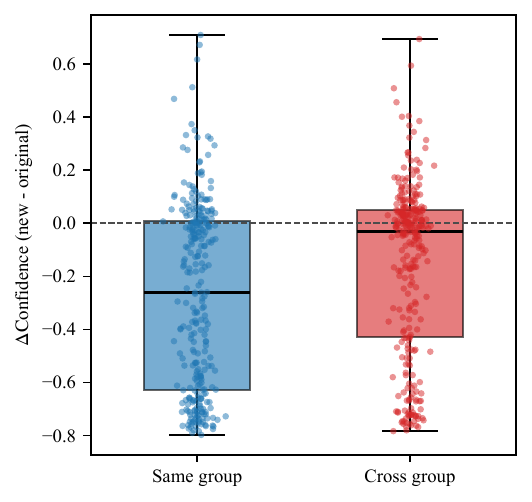}}\\
\subfloat[High-SCR targets]{\includegraphics[width=\columnwidth]{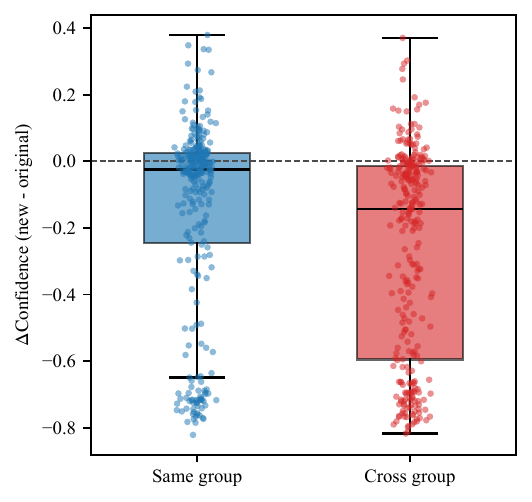}}
\caption{Target transplant on real backgrounds. The vertical axis is the change in confidence (new \(-\) original). Blue boxes show the same-group control, and red boxes show the cross-group transplant. Dashed horizontal line marks zero change. The distribution of individual events is overlaid as jittered points. Events that dropped below the detection threshold after the move are assigned a change of \(-\)original confidence, serving as a conservative lower bound for the net effect.}
\label{fig:transplant}
\end{figure}

We further transplant target pixels onto real backgrounds to isolate the inference-time component of the pairing cost. Targets drawn from the Low-SCR group are pasted at their original coordinates onto High-SCR images containing no ground truth within 8 pixels of the paste location, and vice versa. Same-group transplants serve as controls for the paste boundary. Net effects, computed as cross-group minus same-group, are $+0.1511$ ($p=5.6\times10^{-4}$, $n=44$) for Low$\to$High and $-0.1077$ ($p=0.0229$, $n=36$) for High$\to$Low. The symmetry of the two directions, obtained with real backgrounds and unmodified target pixels, confirms that the detector's confidence depends on target-background pairing. Fig.~\ref{fig:transplant} plots the distribution of these confidence shifts. To rule out censoring bias, a survival analysis excluding missed detections yields the same directional net effects (\(+0.0970\) for Low$\to$High and \(-0.1240\) for High$\to$Low), confirming that the confidence shift is not an artifact of dropped detections.
\begin{figure*}[t]
\centering
\subfloat[NWD metric]{\includegraphics[width=0.48\textwidth]{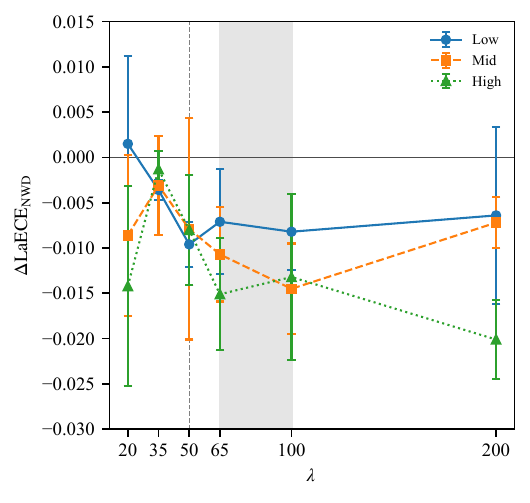}}\hfill
\subfloat[IoU metric]{\includegraphics[width=0.48\textwidth]{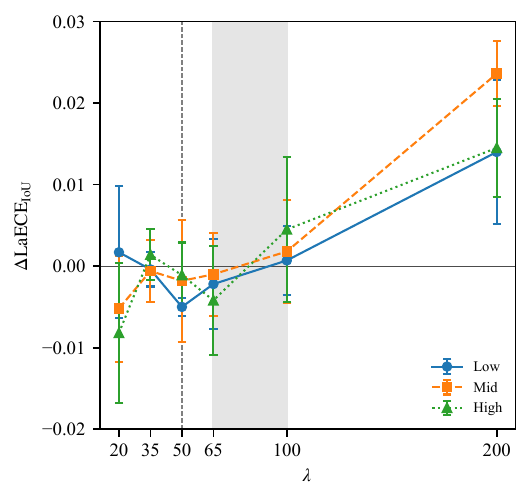}}
\caption{\(\lambda\) sweep over six values, three seeds. Error bars represent seed-level standard deviations of the per-seed difference between the NWD and IoU conditions. (a) \(\Delta\mathrm{LaECE}_{\mathrm{NWD}}\), where negative indicates improvement. (b) \(\Delta\mathrm{LaECE}_{\mathrm{IoU}}\), where positive indicates degradation. The shaded band marks $\lambda\in[65,100]$, where the NWD-metric reductions are largest but the IoU-consistent gain is lost. The vertical dashed line marks the chosen operating point $\lambda=50$, the only setting where the Low-SCR group improves under both metrics with three-seed consistency.}
\label{fig:lambda}
\end{figure*}

\begin{table*}[t]
\centering
\caption{Pairing-broken training under three block sizes at 300 epochs, averaged over three seeds (0/1/42). The $32\times32$ row matches arm~B in Table~\ref{tab:pairing} for LaECE$_{\mathrm{IoU}}$ and mAP. Both LaECE columns are image-level. Corresponding 50-epoch results are reported in the text.}
\label{tab:block}
\setlength{\tabcolsep}{2pt}
\resizebox{\textwidth}{!}{%
\begin{tabular}{lccc|ccc|cc}
\toprule
\multirow{2}{*}{Block} & \multicolumn{3}{c|}{LaECE$_{\text{IoU}}$} & \multicolumn{3}{c|}{LaECE$_{\text{NWD}}$} & \multicolumn{2}{c}{mAP} \\
\cmidrule(lr){2-4}\cmidrule(lr){5-7}\cmidrule(lr){8-9}
 & Low & Mid & High & Low & Mid & High & $_{50}$ & $_{50\text{-}95}$ \\
\midrule
$16\times16$ & $0.0491{\pm}0.0082$ & $0.0488{\pm}0.0053$ & $0.0441{\pm}0.0077$ & $0.0620{\pm}0.0109$ & $0.0654{\pm}0.0065$ & $0.0618{\pm}0.0106$ & $0.9181{\pm}0.0014$ & $0.5062{\pm}0.0043$ \\
$32\times32$ & $0.0494{\pm}0.0054$ & $0.0492{\pm}0.0036$ & $0.0442{\pm}0.0055$ & $0.0624{\pm}0.0080$ & $0.0672{\pm}0.0055$ & $0.0629{\pm}0.0076$ & $0.9253{\pm}0.0062$ & $0.5094{\pm}0.0040$ \\
$64\times64$ & $0.0652{\pm}0.0143$ & $0.0608{\pm}0.0122$ & $0.0561{\pm}0.0105$ & $0.0827{\pm}0.0179$ & $0.0838{\pm}0.0174$ & $0.0791{\pm}0.0136$ & $0.9291{\pm}0.0042$ & $0.5136{\pm}0.0042$ \\
\bottomrule
\end{tabular}}
\end{table*}

The block size controls how much spatial structure survives the remap. Between $16\times16$ and $32\times32$ the pairing cost is flat. The three SCR groups differ by $0.0003$ to $0.0004$ in LaECE$_{\text{IoU}}$, against within-group standard deviations of $0.005$ to $0.008$. The pairing cost saturates across this range, but detection does not: mAP$_{50}$ rises from $0.9181$ at $16\times16$ to $0.9253$ at $32\times32$, a gap of $0.0072$ against a standard deviation of $0.0014$ on the smaller block. The $64\times64$ setting departs from this pattern. LaECE$_{\text{IoU}}$ increases by $0.012$ to $0.016$ across the three groups, because a large block retains most of the within-block thermal structure and the remap no longer destroys the correspondence the analysis is meant to break. The mAP$_{50}$ of $0.9291$ at this size is the highest of the three, but it is obtained under a weaker manipulation. We use $32\times32$ because it sits at the point where the pairing cost has saturated and detection has not yet degraded: a larger block under-manipulates the pairing, a smaller block pays a detection cost without a further increase in pairing cost. 

To verify that the pairing cost is not an artifact of the 300-epoch budget, we repeat the block-remap training under the 50-epoch protocol used in Section~\ref{sec:nwd}. These 50-epoch results are separate from the 300-epoch values in Table~\ref{tab:block}. At 50 epochs, block remap reduces LaECE\(_{\mathrm{IoU}}\) from \(0.0689/0.0649/0.0653\) (the No-BPC row of Table~\ref{tab:main}, trained on the original pairing-preserving set) to \(0.0519/0.0497/0.0491\), and reduces LaECE\(_{\mathrm{NWD}}\) from \(0.0877/0.0895/0.0914\) to \(0.0655/0.0676/0.0689\). The mAP\(_{50}\) cost is \(0.0087\), from \(0.9354\) to \(0.9267\).

Comparing the two budgets, the per-group differences between the 300-epoch condition (Table~\ref{tab:block}, \(32\times32\) row) and the 50-epoch condition are at most \(0.005\) in LaECE\(_{\mathrm{IoU}}\) and below \(0.007\) in LaECE\(_{\mathrm{NWD}}\). The pairing cost persists at matched budget. It is a property of the training distribution, not of training duration.

Block remap is a diagnostic intervention. It requires the training set to be regenerated before retraining, so it does not apply in settings where the training set cannot be altered. It also incurs an mAP\(_{50}\) cost of \(0.0087\). Its role in this paper is to isolate the pairing cost at the data level, complementing the metric-level analysis that follows.

\subsection{Aggregation-Level Bias}
\label{sec:aggregation}
\begin{table}[t]
\centering
\caption{Calibration by SCR group under two aggregation levels, evaluated on the 300-epoch pairing-preserving baseline (Table~\ref{tab:pairing}). The Low-vs-Mid+High effect is \(d=0.217\) under image-level LaECE\(_{\mathrm{IoU}}\) and \(d=0.037\) under the detection-level proxy, a ratio of 5.86.}
\label{tab:aggregation}
\begin{tabular}{lccc}
\toprule
Group & LaECE$_{\text{IoU}}$ (image) & D-ECE (det.) & D-ECE (image proxy) \\
\midrule
Low & 0.0728 & 0.0385 & 0.1185 \\
Mid & 0.0627 & 0.0701 & 0.1166 \\
High & 0.0617 & 0.0739 & 0.1158 \\
\bottomrule
\end{tabular}
\end{table}
Proposition~\ref{prop:dilution} decomposes the aggregation gap into a Between-group term (driven by density ratios \(w_k\)) and a Within-group term (driven by per-image detection-count heterogeneity). Table~\ref{tab:aggregation} instantiates this on the three SCR groups.

The total gap between detection-level and image-level LaECE is \(0.00335\), with the detection-level estimate being larger. Decomposing this gap reveals that the Between-group term contributes \(0.00039\) (11.76\%), while the Within-group term contributes \(0.00296\) (88.24\%). The Within-group term is driven by the Mid and High SCR groups, where a small number of clutter-heavy images have detection counts far exceeding the group average and also carry higher-than-average errors, inflating the detection-level estimate. The Low-SCR group, despite its high total detection count and \(78.6\%\) false-positive rate, contributes negatively to the Within-group term, because its low-confidence false positives depress the detection-level mean below the image-level mean. Image-level aggregation weights each image equally, suppressing the over-representation of clutter-heavy images and restoring the ordering where the Low-SCR group exhibits the highest calibration error (\(0.0728\) vs.\ \(0.0617\)). All subsequent calibration results use image-level aggregation.

\subsection{NWD-Based Calibration}
\label{sec:nwd}
We train BPC variants with IoU and NWD correctness targets under identical assignment, augmentation, and schedule, sweeping $\lambda\in\{20,35,50,65,100,200\}$ on three seeds (0/1/42). The uncalibrated baseline is trained with the standard YOLOv8 script at 50 epochs; we have verified that the BPC training script with \(\lambda=0\) produces identical results, as the calibration loss is disabled before affecting the forward pass.

\begin{table*}[t]
\centering
\caption{Main results at \(\lambda=50\), three seeds. mAP is on the test split (579 images). LaECE values are per-image means averaged over the three seeds. Paired Wilcoxon tests on the per-image means (\(n\) = 121/118/116 unique images per group) yield \(p=5.95\times10^{-8}\), \(8.01\times10^{-8}\), and \(2.07\times10^{-9}\) for the LaECE\(_{\mathrm{NWD}}\) reductions in Low, Mid, and High, respectively. The No-BPC (50 ep) baseline is trained with the standard YOLOv8 script under the identical 50-epoch schedule, augmentation, and matching strategy as the two BPC variants. 
Setting \(\lambda=0\) in the BPC script bypasses the calibration term, making the two conditions structurally equivalent.}
\label{tab:main}
\setlength{\tabcolsep}{3.5pt}
\resizebox{\textwidth}{!}{%
\begin{tabular}{lcccccc}
\toprule
\multirow{2}{*}{Condition} & \multicolumn{2}{c}{mAP} & \multicolumn{2}{c}{LaECE\(_{\mathrm{NWD}}\) (L/M/H)} & \multicolumn{2}{c}{LaECE\(_{\mathrm{IoU}}\) (L/M/H)} \\
\cmidrule(lr){2-3}\cmidrule(lr){4-5}\cmidrule(lr){6-7}
 & \(_{50}\) & \(_{50\text{-}95}\) & Mean & Std & Mean & Std \\
\midrule
No BPC (50 ep) & \(0.9354\pm0.0003\) & \(0.5161\pm0.0028\) & \(0.0877/0.0895/0.0914\) & \(0.0088/0.0075/0.0143\) & \(0.0689/0.0649/0.0653\) & \(0.0073/0.0050/0.0110\) \\
BPC+IoU & \(0.9347\pm0.0030\) & \(0.5166\pm0.0035\) & \(0.0957/0.0993/0.0948\) & \(0.0008/0.0066/0.0035\) & \(0.0761/0.0733/0.0682\) & \(0.0005/0.0043/0.0018\) \\
BPC+NWD (Ours) & \(0.9356\pm0.0018\) & \(0.5138\pm0.0018\) & \(\mathbf{0.0861/0.0914/0.0868}\) & \(\mathbf{0.0020/0.0060/0.0046}\) & \(\mathbf{0.0711/0.0715/0.0671}\) & \(\mathbf{0.0016/0.0037/0.0047}\) \\
\bottomrule
\end{tabular}}
\end{table*}

Relative to BPC+IoU, the NWD target improves LaECE in every SCR group under both anchors. Under LaECE\(_{\mathrm{NWD}}\), the reductions are \(-0.0096\), \(-0.0079\), and \(-0.0080\) in Low, Mid, and High, with \(p<10^{-7}\) in each. The recovery is largest in the low-SCR group, and the High and Mid groups are close to one another. Under LaECE\(_{\mathrm{IoU}}\), the reductions are \(-0.0050\), \(-0.0016\), and \(-0.0011\); the Low group decreases with three-seed consistency (\(p=2.0\times10^{-4}\)). Replacing the IoU target with NWD lowers calibration error in every SCR group, under both anchors and all three seeds.

The comparison against the uncalibrated No-BPC model is more granular and depends on the anchor. Under LaECE\(_{\mathrm{NWD}}\), BPC+NWD is lower than No BPC in the Low and High groups (\(0.0861\) vs.\ \(0.0877\) and \(0.0868\) vs.\ \(0.0914\)) and higher in the Mid group (\(0.0914\) vs.\ \(0.0895\)). Under LaECE\(_{\mathrm{IoU}}\), No BPC is lower than BPC+NWD in every group (\(0.0689/0.0649/0.0653\) vs.\ \(0.0711/0.0715/0.0671\)). The IoU anchor inherits the geometric pathology that this paper identifies: a confidence distribution that matches the IoU variance structure scores well on it, regardless of whether the corresponding detections are physically well localized. The IoU anchor's preference for No BPC is a property of the anchor. The independent center-offset anchor below provides the corresponding check on physical localization.

To eliminate any potential circularity in the evaluation, we introduce a third correctness anchor that is entirely independent of both IoU and NWD: the Euclidean center offset \(\delta_i\) between the predicted box center and its matched ground-truth center. A detection is labeled visually correct if \(\delta_i \le \tau\), with \(\tau \in \{3.0, 4.0\}\) pixels. The results are consistent across both thresholds. In contrast, BPC+NWD recovers most of the localization damage inflicted by BPC+IoU: LaECE\(_{\mathrm{center}}\) falls by 0.0091 in Low, 0.0146 in Mid, and 0.0075 in High relative to BPC+IoU, reaching a level statistically comparable to the uncalibrated baseline (Brier\(_{\mathrm{center}}\) 0.0415 vs.\ 0.0419). The improvement is largest in the mid-SCR group, consistent with the mechanism identified in Section~\ref{sec:correctness_link}. These results provide independent evidence that the NWD target corrects a genuine geometric miscalibration. The same pattern holds at \(\tau=4.0\). 
\begin{table}[t]
\centering
\caption{Calibration under the independent center-offset anchor. A detection is labeled visually correct if the Euclidean center offset \(\delta_i \le \tau\). Results are reported for \(\tau=3.0\) and \(\tau=4.0\) px. Brier\(_{\mathrm{center}}\) is reported for \(\tau=3.0\) only; the pattern at \(\tau=4.0\) is identical (0.0428/0.0459/0.0418 for No-BPC/BPC+IoU/BPC+NWD).}
\label{tab:center}
\setlength{\tabcolsep}{3.5pt}
\resizebox{\columnwidth}{!}{%
\begin{tabular}{lccccccc}
\toprule
\multirow{2}{*}{Condition} & \multicolumn{4}{c}{\(\tau=3.0\) px} & \multicolumn{3}{c}{\(\tau=4.0\) px (LaECE)} \\
\cmidrule(lr){2-5}\cmidrule(lr){6-8}
 & Low & Mid & High & Brier & Low & Mid & High \\
\midrule
No BPC & \textbf{0.1054} & \textbf{0.1233} & 0.1322 & 0.0419 & \textbf{0.1074} & \textbf{0.1252} & 0.1323 \\
BPC+IoU & 0.1204 & 0.1408 & 0.1384 & 0.0455 & 0.1211 & 0.1422 & 0.1382 \\
BPC+NWD (Ours) & 0.1113 & 0.1262 & \textbf{0.1309} & \textbf{0.0415} & 0.1109 & 0.1275 & \textbf{0.1309} \\
\bottomrule
\end{tabular}}
\end{table}

Table~\ref{tab:lambda} lists the full sweep. The two metrics respond differently. Under the NWD-consistent metric, the reductions are generally largest and most consistent in the range \(\lambda\in[65,100]\), with all three SCR groups improving and all seed-level differences sharing the same sign. At \(\lambda=100\), the gains are \(-0.0082/-0.0145/-0.0132\) across Low, Mid, and High. Under the IoU-consistent metric, the response is not monotone. At \(\lambda=50\) the Low group improves with three-seed consistency. The seed-stable IoU-side benefit vanishes beyond \(\lambda=50\). At \(\lambda=200\), the IoU metric degrades in all three groups with consistent seed direction (\(+0.0140/+0.0236/+0.0145\)).
\begin{table*}[t]
\centering
\caption{\(\lambda\) sweep, three seeds (0/1/42). Entries are mean \(\pm\) std of per-seed \(\Delta\), where \(\Delta\) = NWD condition \(-\) IoU condition. Negative \(\Delta\)LaECE$_{\text{NWD}}$ indicates improvement; positive \(\Delta\)LaECE$_{\text{IoU}}$ indicates degradation. Bold marks cells where all three seeds share the same sign. Table~\ref{tab:main} reports the difference of merged absolute values, while this table reports the mean of per-seed differences, resulting in minor rounding discrepancies at \(\lambda=50\).}
\label{tab:lambda}
\setlength{\tabcolsep}{3.5pt}
\resizebox{\textwidth}{!}{%
\begin{tabular}{ccccccc}
\toprule
\multirow{2}{*}{$\lambda$} & \multicolumn{3}{c}{$\Delta$LaECE$_{\text{NWD}}$} & \multicolumn{3}{c}{$\Delta$LaECE$_{\text{IoU}}$} \\
\cmidrule(lr){2-4}\cmidrule(lr){5-7}
 & Low & Mid & High & Low & Mid & High \\
\midrule
20  & $+0.0015\pm0.0097$ & $-0.0086\pm0.0089$ & $\mathbf{-0.0142\pm0.0110}$ & $+0.0017\pm0.0081$ & $-0.0052\pm0.0065$ & $-0.0082\pm0.0086$ \\
35  & $\mathbf{-0.0036\pm0.0011}$ & $-0.0031\pm0.0055$ & $-0.0013\pm0.0020$ & $-0.0004\pm0.0021$ & $-0.0006\pm0.0038$ & $+0.0014\pm0.0031$ \\
50  & $\mathbf{-0.0096\pm0.0025}$ & $-0.0079\pm0.0122$ & $\mathbf{-0.0080\pm0.0061}$ & $\mathbf{-0.0050\pm0.0011}$ & $-0.0018\pm0.0075$ & $-0.0011\pm0.0040$ \\
65  & $\mathbf{-0.0071\pm0.0058}$ & $\mathbf{-0.0107\pm0.0052}$ & $\mathbf{-0.0151\pm0.0062}$ & $-0.0022\pm0.0055$ & $-0.0010\pm0.0051$ & $-0.0042\pm0.0067$ \\
100 & $\mathbf{-0.0082\pm0.0042}$ & $\mathbf{-0.0145\pm0.0050}$ & $\mathbf{-0.0132\pm0.0092}$ & $+0.0007\pm0.0042$ & $+0.0018\pm0.0063$ & $+0.0045\pm0.0089$ \\
200 & $-0.0064\pm0.0098$ & $\mathbf{-0.0072\pm0.0028}$ & $\mathbf{-0.0201\pm0.0044}$ & $\mathbf{+0.0140\pm0.0088}$ & $\mathbf{+0.0236\pm0.0040}$ & $\mathbf{+0.0145\pm0.0060}$ \\
\bottomrule
\end{tabular}}
\end{table*}
The two responses bracket the operating point from different sides. \(\lambda=50\) is the only setting where the Low group shows a decrease under both metrics with consistent seed-level direction. \(\lambda=65\) and \(100\) give larger NWD-metric reductions but lose the IoU-side gain that remains stable across seeds at \(\lambda=50\). Fig.~\ref{fig:lambda} traces this response. We select $\lambda=50$ because it is the only operating point that lowers error under both metric families. This is the conservative choice for deployments where the evaluation metric is not known in advance.

The decomposition also tracks the \(\lambda\) response across the six settings. Under the NWD condition, the geometric distortion \(\mathcal{D}_{\mathrm{geom}}\) is between \(44\%\) and \(52\%\) of the IoU condition at every \(\lambda\). The link-level benefit of the NWD target is therefore a property of the link function itself and does not depend on the calibration weight. What varies with \(\lambda\) is the absolute magnitude. At \(\lambda=200\), every component is higher than at \(\lambda=35\): \(\mathcal{D}_{\mathrm{geom}}\) rises from \(0.0193\) to \(0.0276\), \(\mathcal{E}_{\mathrm{res}}^{\mathrm{det}}\) from \(0.1062\) to \(0.1470\), and \(\mathcal{B}_{\mathrm{agg}}^\ast\) from \(0.0221\) to \(0.0392\). A larger calibration term pushes confidence toward the continuous NWD target and away from the binary anchor, and the resulting residual eventually outweighs the link-level gain. This is the mechanism behind the uniform IoU-side degradation at \(\lambda=200\) reported in Table~\ref{tab:lambda}.

The geometric advantage of NWD does not depend on the precise choice of \(\lambda\). Within \(\lambda\in[35,65]\), the variation in \(\mathcal{D}_{\mathrm{geom}}\) stays within seed-level noise. At \(\lambda=35\) and \(\lambda=50\), \(\mathcal{D}_{\mathrm{geom}}\) is \(0.0193\) and \(0.0205\); the \(0.0012\) difference is within one standard deviation. The choice of \(\lambda=50\) rests on metric consistency. On the IoU side, the improvement at \(\lambda=35\) is \(-0.0004\pm0.0021\) and not distinguishable from zero, whereas at \(\lambda=50\) it is \(-0.0050\pm0.0011\). The sensitive regime is large \(\lambda\), where all three components rise together, not the range \(35\) to \(65\) where the components are stable. The \(\lambda\) analysis therefore delineates an operating interval instead of a single optimum. The full component trajectories are reported in Table~\ref{tab:appendix_lambda_decomp} in the Appendix.

\begin{table}[t]
\centering
\caption{Sensitivity to the NWD scale constant $C$, single seed (42). LaECE is image-level; mAP is on the test split.}
\label{tab:C}
\setlength{\tabcolsep}{4pt}
\begin{tabular}{ccccc}
\toprule
$C$ & LaECE$_{\text{IoU}}$ (L/M/H) & LaECE$_{\text{NWD}}$ (L/M/H) & mAP$_{50}$ & mAP$_{50\text{-}95}$ \\
\midrule
10    & 0.0815/0.0773/0.0740 & 0.0989/0.0996/0.0970 & 0.9364 & 0.5133 \\
15.6  & 0.0725/0.0718/0.0726 & 0.0868/0.0916/0.0915 & 0.9367 & 0.5156 \\
20    & 0.0718/0.0673/0.0692 & 0.0858/0.0847/0.0871 & 0.9316 & 0.5133 \\
25    & 0.0763/0.0705/0.0687 & 0.0915/0.0887/0.0853 & 0.9356 & 0.5107 \\
30    & 0.0807/0.0746/0.0724 & 0.0959/0.0938/0.0900 & 0.9295 & 0.5081 \\
\bottomrule
\end{tabular}
\end{table}
The \(\lambda\) sweep in Table~\ref{tab:lambda} is conducted on three seeds. The $C$ sensitivity analysis in Table~\ref{tab:C} is a single-seed (42) mechanism check to locate the plateau; its conclusion is not used for any statistical comparison across seeds.
The scale constant \(C\) is defined as the mean \(\sqrt{wh}\) of the training-set ground truth (\(C=15.60\) for HIT-UAV). Table~\ref{tab:C} evaluates the sensitivity across \(C\in\{10,15.6,20,25,30\}\). Calibration is best at \(C=20\) (LaECE\(_{\text{IoU}}\) 0.0673--0.0718 against 0.0718--0.0726 at \(C=15.6\)), but mAP\(_{50}\) drops from 0.9367 to 0.9316. At \(C=10\) and \(C=30\) both metrics worsen. The operating value of 15.6 lies on a plateau where both metrics remain stable, confirming that the method does not require precise tuning of \(C\).

\begin{table}[t]
\centering
\caption{Effect of the matching metric. Row A is identical to the NWD-target condition in Table~\ref{tab:main}. Row B replaces IoU matching with NWD matching and keeps the NWD target. All numbers are test-split, three seeds (0/1/42).}
\label{tab:matching}
\setlength{\tabcolsep}{3.5pt}
\resizebox{\columnwidth}{!}{%
\begin{tabular}{lcccccccc}
\toprule
 & \multicolumn{2}{c}{mAP} & \multicolumn{3}{c}{LaECE$_{\text{NWD}}$} & \multicolumn{3}{c}{LaECE$_{\text{IoU}}$} \\
\cmidrule(lr){2-3}\cmidrule(lr){4-6}\cmidrule(lr){7-9}
 & $_{50}$ & $_{50\text{-}95}$ & Low & Mid & High & Low & Mid & High \\
\midrule
A: IoU match + NWD target & $0.9356$ & $0.5138$ & 0.0861 & 0.0914 & 0.0868 & 0.0711 & 0.0715 & 0.0671 \\
B: NWD match + NWD target & $0.9254$ & $0.5028$ & 0.0854 & 0.0804 & 0.0686 & 0.0911 & 0.0972 & 0.0928 \\
\bottomrule
\end{tabular}}
\end{table}

Table~\ref{tab:matching} tests whether NWD should also replace IoU in the matching step. Row A keeps IoU matching and the NWD target; row B replaces the matching metric with NWD and keeps the target. Under the NWD-consistent metric, B improves the Mid and High groups by $0.0107$ and $0.0182$ and leaves the Low group unchanged. Under the IoU-consistent metric, B degrades all three groups by $0.020$ to $0.026$. Detection also drops, by $0.0102$ in mAP$_{50}$ and $0.0110$ in mAP$_{50\text{-}95}$. Replacing IoU matching with NWD matching shifts the calibration bias from the NWD anchor to the IoU anchor and reduces mAP\(_{50}\) by 0.0102. We therefore retain the standard IoU assigner and apply NWD only to the correctness target, which yields gains under both metrics without a matching change.
\begin{table*}[t]
\centering
\caption{Target assignment ablation at \(\lambda=50\) (mean \(\pm\) std over three seeds). NWD-All (NWD target on all matched positives) is the default configuration. NWD Low-IoU Only replaces the target only for matched positives with $\mathrm{IoU}<0.5$; NWD High-IoU Only applies to those with $\mathrm{IoU}\ge 0.5$. The LaECE columns report three-seed statistics. The mAP column reports single-seed values for IoU Baseline, NWD Low-IoU Only, and NWD High-IoU Only, and three-seed means for SizeAdapt and NWD-All. The conclusion of this ablation rests on the calibration evidence. The SizeAdapt rows apply NWD only to matched positives with ground-truth width below \(w_c\); \(w_c=6\) is the critical width from Proposition~\ref{prop:iou_no_floor} and \(w_c=12\) is the median HIT-UAV width.}
\label{tab:assignment}
\setlength{\tabcolsep}{3.5pt}
\resizebox{\textwidth}{!}{%
\begin{tabular}{lccccccc}
\toprule
\multirow{2}{*}{Condition} & \multicolumn{3}{c}{LaECE\(_{\mathrm{NWD}}\)} & \multicolumn{3}{c}{LaECE\(_{\mathrm{IoU}}\)} & \multirow{2}{*}{mAP\(_{50}\)} \\
\cmidrule(lr){2-4}\cmidrule(lr){5-7}
 & Low & Mid & High & Low & Mid & High & \\
\midrule
IoU Baseline & \(0.0957\pm0.0008\) & \(0.0993\pm0.0066\) & \(0.0948\pm0.0035\) & \(0.0761\pm0.0005\) & \(0.0733\pm0.0043\) & \(0.0682\pm0.0018\) & 0.9366 \\
NWD Low-IoU Only & \(0.0925\pm0.0097\) & \(0.0980\pm0.0083\) & \(0.0939\pm0.0089\) & \(0.0735\pm0.0073\) & \(0.0727\pm0.0056\) & \(0.0683\pm0.0077\) & 0.9345 \\
NWD High-IoU Only & \(0.0907\pm0.0061\) & \(0.0895\pm0.0038\) & \(0.0876\pm0.0056\) & \(0.0753\pm0.0049\) & \(0.0695\pm0.0036\) & \(0.0691\pm0.0053\) & 0.9337 \\
SizeAdapt ($w_c{=}6$) & \(0.0937\pm0.0032\) & \(0.1000\pm0.0056\) & \(0.0991\pm0.0081\) & \(0.0742\pm0.0022\) & \(0.0740\pm0.0042\) & \(0.0723\pm0.0062\) & \(0.9326\pm0.0029\) \\
SizeAdapt ($w_c{=}12$) & \(0.0937\pm0.0017\) & \(0.0955\pm0.0023\) & \(0.0928\pm0.0027\) & \(0.0756\pm0.0019\) & \(0.0725\pm0.0023\) & \(0.0706\pm0.0021\) & \(0.9322\pm0.0015\) \\
NWD-All (Ours) & \(\mathbf{0.0861\pm0.0020}\) & \(\mathbf{0.0914\pm0.0060}\) & \(\mathbf{0.0868\pm0.0046}\) & \(\mathbf{0.0711\pm0.0016}\) & \(\mathbf{0.0715\pm0.0037}\) & \(\mathbf{0.0671\pm0.0047}\) & \(\mathbf{0.9356\pm0.0018}\) \\
\bottomrule
\end{tabular}}
\end{table*}

Table~\ref{tab:assignment} isolates the assignment strategy. Replacing only the low-IoU matched positives yields marginal calibration gains but falls far short of the full replacement (e.g., 0.0925 vs. 0.0861 in Low-SCR). Replacing only the high-IoU positives yields partial gains but is notably weaker in the Low-SCR group. Both partial replacements introduce a target inconsistency within the same batch. Under a globally consistent link function, the calibration term retains its Lipschitz property (Section~\ref{sec:correctness_link}). When two link functions are mixed inside one batch, that Lipschitz property no longer holds globally: predictions with nearly identical confidences receive opposing gradient directions depending on whether they fall into the low-IoU or the high-IoU partition. The optimizer faces a target with a discontinuous derivative, and neither partition can be driven toward its own optimum without corrupting the other. Only the full replacement restores a globally consistent, scale-independent correctness target and thereby yields uniformly robust calibration.

The SizeAdapt variants test whether the partition rule is what determines this outcome. They apply NWD only to matched positives whose ground-truth width falls below \(w_c\), with \(w_c=6\) (the critical width of Proposition~\ref{prop:iou_no_floor}) and \(w_c=12\) (the median HIT-UAV width). If the benefit of the full replacement came from finer localization of the geometric pathology, \(w_c=6\) should be the strongest partial variant. It isolates exactly the targets for which IoU collapses. The measurement runs the other way. \(w_c=6\) is the weaker of the two. LaECE\(_{\mathrm{NWD}}\) is \(0.0937/0.1000/0.0991\), against \(0.0937/0.0955/0.0928\) at \(w_c=12\) and \(0.0861/0.0914/0.0868\) for the full replacement. In the Mid group, \(w_c=6\) is at or above the IoU baseline (\(0.1000\) against \(0.0993\)); in the High group, it is worse (\(0.0991\) against \(0.0948\)).

This pattern identifies the source. A partial replacement leaves the batch with two link functions, and adjacent targets on either side of the boundary receive gradients that pull in opposite directions. The sharper the boundary, the closer the mixed targets are in feature space and the larger the conflict. At \(w_c=6\) the boundary cuts through the lower tail of the size distribution; at \(w_c=12\) it cuts through the dense core. The targets on either side of the boundary are nearer to each other at \(w_c=6\) than at \(w_c=12\), and the conflict scales accordingly. The full replacement eliminates the conflict by construction. The partition rule is therefore not what governs the outcome. Any partial replacement reproduces the same failure, and its benefit is consistency of the correctness target rather than finer localization.

\begin{table}[t]
\centering
\caption{Post-hoc calibration at \(\lambda=50\) (three seeds). Left: LaECE\(_{\mathrm{IoU}}\) panel. Right: LaECE\(_{\mathrm{NWD}}\) panel. The fair post-hoc baseline fits Isotonic regression and Temperature Scaling (TS) on the validation split only, both with IoU as the target. Both mappings are then applied to the test set without any test-time ground truth. The Oracle row is fitted on the test set with IoU as the target and serves as an upper bound. The last row applies an isotonic map fitted on the BPC+NWD model's own validation split to its test confidences, and reports three-decimal values. Brier\(_{\mathrm{IoU}}\) is reported under two conventions: TP-only when comparing post-hoc maps on correctly localized detections, and full-detection when comparing the composite NWD+isotonic configuration against the other methods.}
\label{tab:posthoc}
\setlength{\tabcolsep}{3pt}
\resizebox{\columnwidth}{!}{%
\begin{tabular}{lcccccc}
\toprule
\multirow{2}{*}{Method} & \multicolumn{3}{c}{LaECE\(_{\mathrm{IoU}}\) (independent anchor)} & \multicolumn{3}{c}{LaECE\(_{\mathrm{NWD}}\) (aligned anchor)} \\
\cmidrule(lr){2-4}\cmidrule(lr){5-7}
 & Low & Mid & High & Low & Mid & High \\
\midrule
BPC+IoU (no post-proc.) & 0.0761 & 0.0733 & 0.0682 & 0.0957 & 0.0993 & 0.0948 \\
+ Temperature Scaling & 0.0718 & 0.0765 & 0.0831 & 0.0683 & 0.0583 & 0.0556 \\
+ Fair Post-hoc (Iso on Val) & 0.0671 & 0.0635 & 0.0611 & 0.0874 & \textbf{0.0891} & 0.0868 \\
+ Oracle Isotonic (test-fit) & 0.0661 & 0.0623 & 0.0602 & 0.0867 & 0.0870 & 0.0845 \\
+ NWD (Ours, end-to-end) & 0.0711 & 0.0715 & 0.0671 & \textbf{0.0861} & 0.0914 & \textbf{0.0868} \\
\midrule
BPC+NWD + Iso (val-fit) & 0.064 & 0.064 & 0.061 & 0.082 & 0.088 & 0.086 \\
\bottomrule
\end{tabular}}
\end{table}

The decomposition in Table~\ref{tab:component_decomp} explains the pattern: temperature scaling and isotonic regression act on the confidence mapping, and NWD acts on the link function. On the IoU anchor, the fair isotonic baseline achieves the lowest mean error because its non-parametric map is fitted to the empirical IoU marginal. This is alignment to the evaluation anchor, not correction of the IoU geometry that suppresses visually correct tiny detections. Temperature scaling helps only in the low-SCR group on this anchor (0.0718 against 0.0761) and increases LaECE\(_{\mathrm{IoU}}\) in the Mid and High groups (0.0765 against 0.0733, 0.0831 against 0.0682).

The independent anchors separate the two mechanisms. On LaECE\(_{\mathrm{NWD}}\), temperature scaling is lower than BPC+NWD in all three groups (0.0683/0.0583/0.0556 vs.\ 0.0861/0.0914/0.0868), but its TP-only Brier\(_{\mathrm{IoU}}\) remains at the uncalibrated level (0.02906 vs.\ 0.02907; BPC+NWD: 0.02680). Its LaECE\(_{\mathrm{NWD}}\) gain therefore reflects a global shift of the confidence distribution; alignment with correctness is unchanged. On the center-offset anchor, BPC+NWD reaches a Brier\(_{\mathrm{center}}\) of 0.0415 against 0.0455 for BPC+IoU and 0.0419 for the uncalibrated model. Against fair isotonic on LaECE\(_{\mathrm{NWD}}\), BPC+NWD is slightly lower in the low-SCR group (0.0861 vs.\ 0.0874), identical in the high-SCR group (0.0868), and slightly higher in the mid-SCR group (0.0914 vs.\ 0.0891); these differences are small and should be read as comparability, not dominance. The decisive separation is that isotonic fits the IoU marginal and leaves \(\mathcal{D}_{\mathrm{geom}}\) unchanged, whereas NWD lowers \(\mathcal{D}_{\mathrm{geom}}\) and improves independent proper scores and the physical center-offset anchor.

The complementarity of the two corrections is internal to the correctness target, unlike the two error sources in Sections~\ref{sec:pairing} and~\ref{sec:aggregation}, which lie outside the metric itself. An isotonic map fitted on the BPC+NWD model's own validation split lowers LaECE\(_{\mathrm{IoU}}\) from 0.0711/0.0715/0.0671 to 0.064/0.064/0.061 and LaECE\(_{\mathrm{NWD}}\) from 0.0861/0.0914/0.0868 to 0.082/0.088/0.086. Against IoU+isotonic, the combination is comparable on LaECE\(_{\mathrm{IoU}}\), with a 0.0005 gap at Mid favoring IoU+isotonic, and lower on LaECE\(_{\mathrm{NWD}}\) in all three groups. The image-level Brier\(_{\mathrm{IoU}}\) also drops from 0.0200 to 0.0179 (\(p=2.3\times10^{-16}\), three of three seeds), but the Brier comparison against IoU+isotonic is not stable: the per-seed difference flips sign, and the effect sizes are at or below 0.002. Brier does not rank the two composite configurations in this setting. The two corrections are complementary and do not substitute for one another.

The independent value of NWD is separate from the combination. The isotonic mapping is a fixed lookup table estimated on the validation split and carried in the deployed graph; it must be refitted whenever the sensor, altitude, or scene composition changes. NWD is applied during training and leaves the inference graph unchanged. In deployments where the evaluation distribution is unknown or unstable, NWD provides calibrated confidence without a data-side refit. A post-hoc map requires a representative validation set at deployment time.

\begin{table}[t]
\centering
\caption{Effect of target distribution matching at \(\lambda=50\) (mean \(\pm\) std over three seeds). NWD-Matched rescales NWD targets to have the same mean and variance as the IoU targets.}
\label{tab:target_match}
\setlength{\tabcolsep}{3.5pt}
\resizebox{\columnwidth}{!}{%
\begin{tabular}{lcccccc}
\toprule
\multirow{2}{*}{Condition} & \multicolumn{3}{c}{LaECE\(_{\mathrm{NWD}}\)} & \multicolumn{2}{c}{mAP} \\
\cmidrule(lr){2-4}\cmidrule(lr){5-6}
 & Low & Mid & High & \(_{50}\) & \(_{50\text{-}95}\) \\
\midrule
BPC+IoU & \(0.0957\pm0.0008\) & \(0.0993\pm0.0066\) & \(0.0948\pm0.0035\) & 0.9366 & 0.5169 \\
BPC+NWD (Ours) & \(\mathbf{0.0861\pm0.0020}\) & \(\mathbf{0.0914\pm0.0060}\) & \(\mathbf{0.0868\pm0.0046}\) & \textbf{0.9367} & \textbf{0.5156} \\
NWD-Matched & \(0.0974\pm0.0046\) & \(0.0937\pm0.0045\) & \(0.0923\pm0.0063\) & 0.9287 & 0.5107 \\
\bottomrule
\end{tabular}}
\end{table}

Forcing the NWD targets to match the mean and variance of the IoU targets degrades both calibration and mAP (Table~\ref{tab:target_match}). The gain therefore comes from the geometric ordering of the NWD target, rather than from a shift in the target distribution.

%
\begin{figure}[!t]
\centering
\includegraphics[width=\columnwidth]{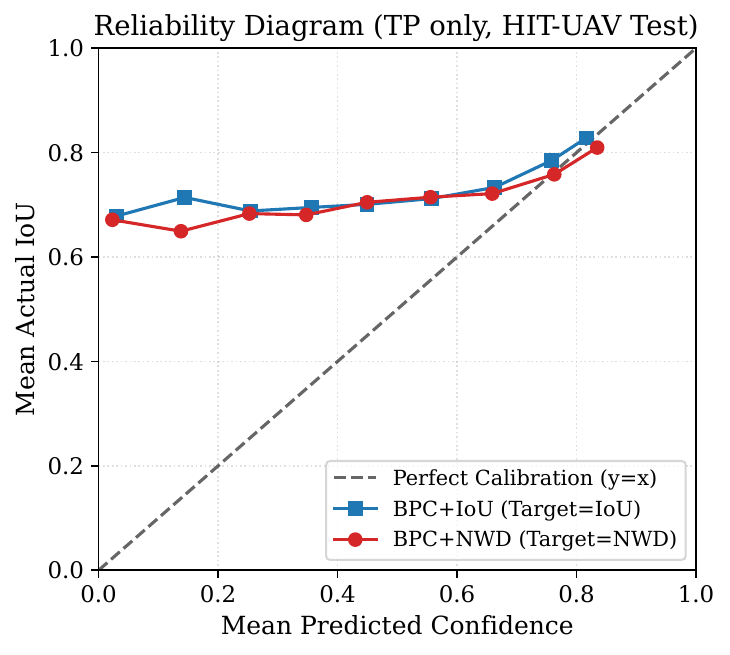}
\caption{Reliability diagrams for the IoU-trained and NWD-trained models on the HIT-UAV test split. The diagonal represents perfect calibration. Both models track the diagonal for true positives, while the NWD target yields a tighter alignment in mid-confidence bins (computed on seed 42 for visualization).}
\label{fig:reliability}
\end{figure}

Fig.~\ref{fig:reliability} compares the reliability diagrams of the IoU and NWD conditions. Both models track the perfect calibration diagonal closely for true positives. The NWD condition exhibits tighter adherence to the diagonal in the mid-confidence bins (0.4--0.6), aligning with its superior LaECE scores in Table~\ref{tab:main}.

\begin{table}[t]
\centering
\caption{LaECE\(_{\mathrm{IoU}}\) breakdown by target size and SCR group at \(\lambda=50\) (single seed 42). $\Delta$ = NWD $-$ IoU (negative indicates improvement).}
\label{tab:scale_scr}
\setlength{\tabcolsep}{3.5pt}
\resizebox{\columnwidth}{!}{%
\begin{tabular}{clcccc}
\toprule
\multirow{2}{*}{SCR} & \multirow{2}{*}{Size (px\(^2\))} & \multirow{2}{*}{BPC+IoU} & \multirow{2}{*}{BPC+NWD} & \multicolumn{1}{c}{$\Delta$} \\
\cmidrule(lr){5-5}
 & & & & (NWD \(-\) IoU) \\
\midrule
\multirow{3}{*}{High} & $<200$ & 0.1137 & 0.0988 & $\mathbf{-0.0149}$ \\
 & $200$--$500$ & 0.0870 & 0.1021 & $+0.0151$ \\
 & $>500$ & 0.1343 & 0.1335 & $-0.0008$ \\
\midrule
\multirow{3}{*}{Mid} & $<200$ & 0.1348 & 0.1228 & $\mathbf{-0.0120}$ \\
 & $200$--$500$ & 0.1055 & 0.0990 & $-0.0065$ \\
 & $>500$ & 0.1166 & 0.1007 & $\mathbf{-0.0159}$ \\
\midrule
\multirow{3}{*}{Low} & $<200$ & 0.1696 & 0.1543 & $\mathbf{-0.0153}$ \\
 & $200$--$500$ & 0.1949 & 0.2028 & $+0.0080$ \\
 & $>500$ & 0.2068 & 0.1997 & $-0.0070$ \\
\bottomrule
\end{tabular}}
\end{table}

Table~\ref{tab:scale_scr} decouples target scale from background complexity. In the HIT-UAV dataset, small targets naturally co-occur with low SCR, which makes SCR a strong macro-level proxy for scale. However, when stratifying by size directly, we observe that in the smallest size bin (\(<200\text{ px}^2\)), the NWD target yields a consistent calibration improvement across \emph{all} three SCR groups (\(\Delta = -0.0149, -0.0120, -0.0153\) for High, Mid, and Low). For larger targets, the improvement diminishes or fluctuates regardless of the background. The degradation in the 200--500 px² bins is consistent with the scale-heterogeneity caveat reported for NWD in assignment~\cite{coswa}: for targets where IoU already provides an informative localization gradient, the smooth Gaussian modeling of NWD weakens that signal. This confirms target scale as the fundamental geometric driver of the NWD gain, while background complexity modulates its macroscopic severity.

\begin{table}[t]
\centering
\caption{Calibration error stratified by IoU threshold on the test split. $\Delta$ = NWD $-$ IoU.}
\label{tab:iou_subset}
\begin{tabular}{lccc}
\toprule
Subset & BPC+IoU & BPC+NWD & $\Delta$ (NWD \(-\) IoU) \\
\midrule
IoU \(<\) 0.5 & 0.0634 & 0.0695 & $+0.0061$ \\
IoU \(\ge\) 0.5 & 0.1915 & 0.1554 & \(\mathbf{-0.0361}\) \\
\bottomrule
\end{tabular}
\end{table}

Table~\ref{tab:iou_subset} stratifies the calibration error by the IoU threshold. On the subset with IoU \(\ge 0.5\), the NWD target drastically reduces the calibration error (0.1915 to 0.1554, \(\Delta = -0.0361\)), as IoU cannot distinguish between a barely acceptable box and a tightly localized one. 

On the IoU \(<\) 0.5 subset, the aggregate error slightly increases from 0.0634 to 0.0695. Under the continuous soft target, their \(y_i\) is the matched IoU value (e.g., 0.3), not zero. The NWD target assigns a higher value to these visually correct detections, intentionally elevating their confidence. This elevation increases the absolute deviation \(|y_i - p_i|\) when evaluated under the IoU target \(y_i\), but it is the intended correction of the under-confidence caused by the IoU pathology. The global calibration improves, as confirmed by the overall LaECE and the independent center-offset anchor in Table~\ref{tab:center}.

\begin{figure}[!t]
\centering
\includegraphics[width=\columnwidth]{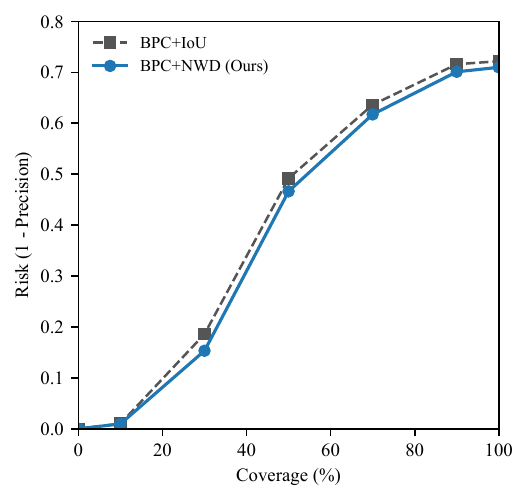}
\caption{Risk-coverage curves for selective prediction on the HIT-UAV test split. At any given coverage level, the NWD-calibrated model yields a lower risk (fewer false alarms) than the IoU baseline. The risk reduction is most pronounced at 30\% coverage, where the risk drops from 0.1861 to 0.1533 (single seed 42).}
\label{fig:risk_coverage}
\end{figure}

\begin{figure}[!t]
\centering
\includegraphics[width=\columnwidth]{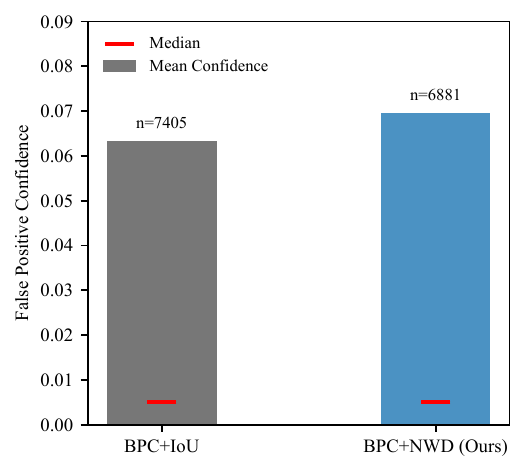}
\caption{False positive confidence distribution. The NWD model yields a similar median FP confidence (0.0051) and a slightly higher mean, but produces fewer total false positives (6,881 vs.\ 7,405). Calibration improvements are not achieved by inflating the confidence of incorrect predictions (single seed 42).}
\label{fig:fp_confidence}
\end{figure}

Finally, we evaluate the downstream utility of the calibrated models via selective prediction and false-positive behavior (Fig.~\ref{fig:risk_coverage} and Fig.~\ref{fig:fp_confidence}). 
The NWD model consistently achieves lower risk across all coverage levels, with the largest margin at 30\% coverage (Risk 0.1533 vs.\ 0.1861). This indicates that when an operator retains only the top 30\% most confident detections, the NWD model yields significantly fewer false alarms. 
This gain is not achieved by indiscriminately inflating confidence. As shown in Fig.~\ref{fig:fp_confidence}, the median false-positive confidence remains identical for both models (0.0051), while the total FP count drops from 7,405 to 6,881. The calibration improvement thus translates to genuine decision-making reliability.

\subsection{Cross-Dataset Validation on FLIR ADAS}
\label{sec:flir}

To test whether the observed mechanism transfers beyond the aerial perspective, we repeat the evaluation on the FLIR ADAS dataset using thermal-only input. Targets here are captured at street level and occupy a much larger pixel area. The baseline detector is trained with \(\lambda=0\) under the same 50-epoch schedule as the BPC variants, and achieves a test mAP\(_{50}\) of 0.7405 and mAP\(_{50\text{-}95}\) of 0.4373 (three seeds). The NWD scale constant is set to \(C=17.65\) pixels, the mean \(\sqrt{wh}\) of the FLIR training set. SCR tertile boundaries are 0.5943 and 1.2329. We use a detection threshold of 0.001, and FP is defined as any detection with \(\mathrm{IoU}<0.5\) against all ground truths. The high-SCR group averages 11 detections per image, dominating the FP volume.

\begin{table}[t]
\centering
\caption{FLIR ADAS main results at \(\lambda=50\) (three seeds). mAP is on the test split (3,749 images). \(\Delta\) = NWD \(-\) IoU.}
\label{tab:flir_main}
\setlength{\tabcolsep}{3.5pt}
\resizebox{\columnwidth}{!}{%
\begin{tabular}{lcccccccc}
\toprule
 & \multicolumn{2}{c}{mAP} & \multicolumn{3}{c}{LaECE\(_{\mathrm{NWD}}\)} & \multicolumn{3}{c}{LaECE\(_{\mathrm{IoU}}\)} \\
\cmidrule(lr){2-3}\cmidrule(lr){4-6}\cmidrule(lr){7-9}
Condition & \(_{50}\) & \(_{50\text{-}95}\) & Low & Mid & High & Low & Mid & High \\
\midrule
No BPC & 0.7405 & 0.4373 & 0.0886 & 0.0786 & 0.0836 & 0.0705 & 0.0620 & 0.0657 \\
BPC+IoU & 0.7326 & 0.4311 & 0.1005 & 0.0774 & 0.0798 & 0.0816 & 0.0610 & 0.0622 \\
BPC+NWD & 0.7225 & 0.4263 & 0.0753 & 0.0784 & 0.0796 & 0.0637 & 0.0633 & 0.0628 \\
\midrule
\(\Delta\) & \(-0.0101\) & \(-0.0048\) & \(-0.0252\) & \(+0.0010\) & \(-0.0002\) & \(-0.0179\) & \(+0.0023\) & \(+0.0006\) \\
\bottomrule
\end{tabular}}
\end{table}

In the mid-SCR group, the NWD target is slightly worse in both metrics ($+0.0010$ in LaECE\(_{\mathrm{NWD}}\) and $+0.0023$ in LaECE\(_{\mathrm{IoU}}\)), and all three seeds share this direction. In the high-SCR group, the LaECE\(_{\mathrm{NWD}}\) difference is negligible and slightly negative ($-0.0002$, i.e., a marginal improvement), while LaECE\(_{\mathrm{IoU}}\) is slightly worse ($+0.0006$) with mixed seed directions. These deviations precisely corroborate Proposition~\ref{prop:iou_no_floor}. On FLIR, where targets are larger and pixel jitter is less destructive, IoU already provides a reliable gradient. The neutral-to-slightly-negative response on Mid-SCR targets exactly delineates the operating boundary of the method: NWD is irreplaceable for tiny targets but incurs a small mAP cost where the IoU metric is already geometrically sufficient.

This cross-view thermal-infrared validation maps out the operating regime. On street-level imagery, larger targets are less susceptible to the pixel-level jitter that drives the IoU pathology (Proposition~\ref{prop:iou_no_floor}). The calibration gain of NWD is concentrated in the low-SCR regime, where a few pixels of displacement dominate the overlap ratio, while the mid and high-SCR groups remain neutral. This pattern matches the mechanism identified on HIT-UAV. In Figure~\ref{fig:scatter}(c), 90.5\% of the detections that IoU labels incorrect are judged correct by NWD. The mAP cost on FLIR ($-0.0101$ at mAP\(_{50}\)) is larger than on HIT-UAV, where the NWD target is mAP-neutral. The difference reflects the target-scale distribution: on HIT-UAV, nearly all targets fall in the tiny-object regime where IoU is pathological, so replacing IoU with NWD is unambiguously beneficial; on FLIR, a substantial fraction of targets occupy a scale at which IoU is already informative, and the NWD target provides a weaker localization signal for those detections.

The FLIR dataset provides an independent confirmation of the aggregation bias analysed in Proposition~\ref{prop:dilution}. As shown in Table~\ref{tab:flir_agg}, detection-level D-ECE ranks the high-SCR group as the best calibrated (0.0504) and the low-SCR group as the worst (0.0648). Image-level LaECE\(_{\text{IoU}}\) reverses this ordering entirely (0.0692 for high-SCR and 0.0580 for low-SCR). The high-SCR group contains 35,351 false positives against 6,695 true positives. This reversal reproduces the empirical dominance reported in Corollary~\ref{cor:direction} for HIT-UAV, although the group ordering differs because FLIR's high-SCR group contains the largest false-positive volume. The same mechanism applies: detection-level aggregation introduces a sign-varying bias whenever a group's per-image detection count is highly heterogeneous. In FLIR's high-SCR group the bias is negative, because the sheer volume of low-confidence false positives drags the detection-level estimate toward zero.
\begin{table}[t]
\centering
\caption{FLIR ADAS calibration under two aggregation levels, evaluated on the 300-epoch baseline (early-stopped at 142 epochs, single seed 42). Detection-level D-ECE and image-level LaECE\(_{\text{IoU}}\) produce opposite rankings across SCR groups.}
\label{tab:flir_agg}
\setlength{\tabcolsep}{4pt}
\begin{tabular}{lccc}
\toprule
Group & LaECE\(_{\text{IoU}}\) (image) & D-ECE (det.) \\
\midrule
High & 0.0692 & 0.0504 \\
Mid  & 0.0658 & 0.0558 \\
Low  & 0.0580 & 0.0648 \\
\bottomrule
\end{tabular}
\end{table}
\section{Discussion}

The calibration error in an ordinary detector originates from three independent sources: the geometric scale mismatch (the IoU pathology), the aggregation protocol, and the residual model quality. In UAV imagery, this geometric flaw is amplified by complex thermal backgrounds. The low-SCR group carries the largest absolute error. By decoupling scale and SCR, we demonstrate that the NWD target addresses this root cause. This error is further distorted by detection-level aggregation when per-image detection counts differ, with the direction depending on whether the group's detection-level mean error exceeds or falls below its image-level mean error. Correcting the aggregation and the metric reveals the residual error, which is recoverable through the NWD target without compromising detection accuracy.

NWD assigns larger target values to low-IoU detections, yielding higher predicted confidences that align with visual correctness. The NWD condition reduces the localization-aware calibration error under both the IoU and NWD evaluation metrics (Table~\ref{tab:main}). Table~\ref{tab:matching} shows that NWD must act strictly as a soft target corrector on the existing IoU assignment; replacing the assigner itself with NWD matching degrades both mAP and IoU-consistent calibration. 
While the NWD target distribution is inherently shifted upward relative to IoU, Table~\ref{tab:target_match} demonstrates that forcing a distribution match degrades performance. The effectiveness of NWD stems from its geometric ordering, which is scale-invariant for a fixed scale constant \(C\).

A deployment consideration follows from where the calibration term acts. Post-hoc calibration methods, while effective in offline benchmarks, require a separate fitting phase on ground-truth data and carry the risk of distribution shift when deployed in novel environments. The NWD target is confined to training, leaving the inference graph structurally identical to a detector trained without calibration. The detector can be exported, quantized, or compiled for an embedded target without carrying a calibration head. The calibration strength \(\lambda\) is fixed at training and does not appear in the exported model. This training-time formulation avoids the coupling between deployment and offline fitting, making it particularly suitable for UAV embedded systems where memory and latency budgets are tight.

The $\lambda$ response is not monotone. The NWD-metric optimum is broad and sits around $\lambda=65$--$100$, where all three SCR groups improve with consistent seed direction. The IoU metric tolerates low \(\lambda\). However, the three-seed consistency on the IoU side breaks down past \(\lambda=50\), and by \(\lambda=200\) the degradation becomes unanimous across all groups. The two constraints overlap near $\lambda=50$, which is where the Low group improves under both metrics with three-seed consistency. That overlap is narrow: $\lambda=65$ already gives larger NWD-metric reductions but loses the IoU-side improvement, and $\lambda=200$ trades a further NWD gain in the High group for consistent IoU degradation in all groups. A single global $\lambda$ therefore cannot satisfy every SCR group simultaneously, and the choice of 50 privileges the Low group's dual-metric improvement over the larger NWD-only gains available at higher $\lambda$.

The pairing cost reported in Section~\ref{sec:pairing} and the metric flaw addressed by the NWD target are two orthogonal error sources. NWD addresses the metric flaw. The pairing cost is a \emph{distribution-level} effect: it originates in $P(y=1\mid x_t,x_b)$ deviating from $P(y=1\mid x_t)$ whenever the background carries label-relevant thermal structure. The metric flaw is a \emph{geometric-level} effect: it originates in the IoU link function collapsing under pixel-level jitter, independent of what the background contains. Because an intervention on one does not address the other, the two sources can be diagnosed and corrected independently (Sections~\ref{sec:pairing} and \ref{sec:correctness_link}). NWD redefines the correctness target and therefore targets the geometric source. Reducing the pairing cost requires distribution-level intervention, such as paired thermal augmentation or explicit pairing regularization. A joint intervention that simultaneously reduces both is left for future work. The block-remap intervention at 50 epochs (Section~\ref{sec:pairing}) occupies a different operating point. It reduces LaECE at a measurable mAP cost of \(0.0087\), whereas the metric-level NWD formulation is mAP-neutral. The two interventions are complementary: block remap acts on the data distribution, and NWD acts on the correctness target.

The NWD target is mAP-neutral on HIT-UAV, where nearly all targets fall in the tiny-object regime and IoU is pathological. On FLIR ADAS, where a substantial fraction of targets are large enough for IoU to be informative, the NWD target introduces a \(-0.0101\) mAP\(_{50}\) cost. The gap tracks how much of each dataset falls in the tiny-object regime: the smooth Gaussian modeling of NWD weakens the localization signal for detections where IoU is already reliable. Consequently, the NWD target is most appropriate for deployment scenarios dominated by tiny targets.

\textbf{Multi-anchor evaluation framework.} A methodological contribution of this work is the evaluation under three distinct correctness anchors. The theoretical bound establishes a mathematical coupling under \(y^\ast\). The empirical validation relies on independent geometric properties. The IoU anchor reflects the regression-derived criterion, the NWD anchor captures continuous geometric alignment, and the center-offset anchor measures physical localization. All three are reported jointly (Table~\ref{tab:center}). This multidimensional protocol confirms that the NWD training target corrects the geometric miscalibration, rather than merely fitting a specific metric.

\section{Conclusion}

For UAV infrared person detection, IoU-dominated correctness targets are geometrically ill-posed for tiny objects. We show that this pathology propagates into calibration, and that a continuous NWD target reduces calibration error across all SCR groups without sacrificing mAP on HIT-UAV. We further isolate two error sources outside the metric, thermal pairing and aggregation bias, and demonstrate that the NWD target and post-hoc isotonic calibration are complementary. The decomposition of Section~\ref{sec:unified_bound} separates the sources that a metric-level correction can address from those that require intervention outside the metric.

\appendices
\section{Component Decomposition Across the \(\lambda\) Sweep}
\label{app:lambda_decomp}

The three components of Theorem~\ref{thm:unified_bound} at six calibration weights, for both BPC+IoU and BPC+NWD, are in Table~\ref{tab:appendix_lambda_decomp}. All values are three-seed mean \(\pm\) standard deviation on the HIT-UAV test split.

\begin{table}[h]
\centering
\caption{Decomposition of the calibration error across the \(\lambda\) sweep (three seeds, mean \(\pm\) std). \(\mathcal{E}_{\mathrm{MAD}}\) is omitted because it is computed with each condition's own \(L\) and is not comparable across rows (see Table~\ref{tab:component_decomp}).}
\label{tab:appendix_lambda_decomp}
\setlength{\tabcolsep}{3.5pt}
\resizebox{\columnwidth}{!}{%
\begin{tabular}{clccc}
\toprule
\(\lambda\) & Condition & \(\mathcal{D}_{\mathrm{geom}}\) & \(\mathcal{E}_{\mathrm{res}}^{\mathrm{det}}\) & \(\mathcal{B}_{\mathrm{agg}}^\ast\) \\
\midrule
\multirow{2}{*}{20} 
 & BPC+IoU & \(0.0388\pm0.0024\) & \(0.1090\pm0.0027\) & \(+0.0220\pm0.0030\) \\
 & BPC+NWD & \(0.0171\pm0.0020\) & \(0.1024\pm0.0090\) & \(+0.0239\pm0.0029\) \\
\midrule
\multirow{2}{*}{35} 
 & BPC+IoU & \(0.0368\pm0.0017\) & \(0.1065\pm0.0042\) & \(+0.0246\pm0.0031\) \\
 & BPC+NWD & \(0.0193\pm0.0022\) & \(0.1062\pm0.0062\) & \(+0.0221\pm0.0002\) \\
\midrule
\multirow{2}{*}{50} 
 & BPC+IoU & \(0.0432\pm0.0050\) & \(0.1168\pm0.0046\) & \(+0.0257\pm0.0022\) \\
 & BPC+NWD & \(0.0205\pm0.0022\) & \(0.1087\pm0.0040\) & \(+0.0218\pm0.0019\) \\
\midrule
\multirow{2}{*}{65} 
 & BPC+IoU & \(0.0440\pm0.0046\) & \(0.1203\pm0.0040\) & \(+0.0252\pm0.0018\) \\
 & BPC+NWD & \(0.0210\pm0.0037\) & \(0.1110\pm0.0088\) & \(+0.0254\pm0.0009\) \\
\midrule
\multirow{2}{*}{100} 
 & BPC+IoU & \(0.0463\pm0.0003\) & \(0.1294\pm0.0027\) & \(+0.0290\pm0.0021\) \\
 & BPC+NWD & \(0.0225\pm0.0016\) & \(0.1179\pm0.0049\) & \(+0.0282\pm0.0004\) \\
\midrule
\multirow{2}{*}{200} 
 & BPC+IoU & \(0.0545\pm0.0050\) & \(0.1548\pm0.0077\) & \(+0.0374\pm0.0045\) \\
 & BPC+NWD & \(0.0276\pm0.0023\) & \(0.1470\pm0.0044\) & \(+0.0392\pm0.0015\) \\
\bottomrule
\end{tabular}}
\end{table}
\section{Image-Level Brier Under Two Composite Configurations}
\label{app:brier_combination}

Table~\ref{tab:appendix_combination} reports the image-level Brier\(_{\mathrm{IoU}}\) of the two composite configurations on the images common to all three seeds, stratified by per-image detection count. The per-seed difference between the two configurations changes sign, and the effect sizes are small in absolute terms. The stratified rows do not show a stable direction: the 6--20 bin favors NWD+Iso (\(p=0.0098\)), while the sparsest bin and two of the three per-seed values in the main table favor IoU+Iso. The table is provided for completeness; the main comparison in the paper uses LaECE and the center-offset anchor, where the direction is consistent across seeds.

\begin{table}[h]
\centering
\caption{Image-level Brier\(_{\mathrm{IoU}}\) of the two composite configurations on common images. The first four rows report per-seed and three-seed combined values; the last four rows stratify the combined set by per-image detection count. The per-seed \(n\) is the number of images detected by both models in that seed; the combined \(n=490\) is the number of images detected by both models in all three seeds.}
\label{tab:appendix_combination}
\setlength{\tabcolsep}{3.5pt}
\resizebox{\columnwidth}{!}{%
\begin{tabular}{lcccccc}
\toprule
Group & \(n\) & NWD+Iso & IoU+Iso & \(\Delta\) & \(p\) \\
\midrule
seed 0 & 533 & 0.0188 & 0.0175 & \(+0.00127\) & 0.0127 \\
seed 1 & 537 & 0.0177 & 0.0155 & \(+0.00217\) & 0.685 \\
seed 42 & 512 & 0.0184 & 0.0186 & \(-0.00019\) & 0.0026 \\
3-seed combined & 490 & 0.0197 & 0.0185 & \(+0.00119\) & 0.032 \\
\midrule
\(\le 5\) & 88 & 0.0255 & 0.0160 & \(+0.00954\) & 0.092 \\
6--20 & 216 & 0.0175 & 0.0184 & \(-0.00087\) & 0.0098 \\
21--50 & 117 & 0.0177 & 0.0176 & \(+0.00014\) & 0.68 \\
51--200 & 69 & 0.0223 & 0.0235 & \(-0.00123\) & 0.196 \\
\bottomrule
\end{tabular}}
\end{table}

\end{document}